\documentclass{article}
\usepackage{tabularx}
\newcolumntype{Y}{>{\centering\arraybackslash}X}
\usepackage{PRIMEarxiv}
\usepackage[utf8]{inputenc}
\usepackage[T1]{fontenc}
\usepackage{url}
\usepackage{booktabs}
\usepackage{amsmath,amssymb,amsfonts}
\usepackage{microtype}
\usepackage{graphicx}
\graphicspath{{./}{./figures/}{figures/}}
\usepackage{multirow}
\usepackage{natbib}
\usepackage{array}
\usepackage[table]{xcolor}
\usepackage{algorithm}
\usepackage{algpseudocode}
\usepackage{xspace}
\usepackage{enumitem}
\usepackage{tabularx}
\usepackage{makecell}
\usepackage{float}
\usepackage{placeins}

\usepackage[bookmarks=true,colorlinks=true]{hyperref}
\hypersetup{
  pdftitle  ={AlignBeam: Inference-Time Alignment Transfer via Cross-Vocabulary Logit Mixing},
  pdfauthor ={Chirag Chawla, Pratinav Seth, Vinay Kumar Sankarapu},
  pdfsubject={Inference-time safety alignment for domain-specialist LLMs via cross-vocabulary logit mixing},
  pdfkeywords={LLM safety, inference-time alignment, logit mixing, cross-vocabulary, domain fine-tuning, SafeDecoding, beam search}
}

\definecolor{gainHigh}{HTML}{A8D5A2}   
\definecolor{gainMed}{HTML}{C8E6C9}    
\definecolor{gainLow}{HTML}{E8F5E9}    
\definecolor{gainNeg}{HTML}{FFCDD2}    
\definecolor{pendingCell}{HTML}{FFF9C4} 
\definecolor{rowOurs}{HTML}{F1F8E9}    

\newcommand{\alignbeam}{\textsc{AlignBeam}\xspace}
\newcommand{\lbm}{\textsc{LBM}\xspace}
\newcommand{\lbmdrop}{\textsc{LBM-Drop}\xspace}
\newcommand{\lbmfirst}{\textsc{LBM-First}\xspace}
\newcommand{\lbmexact}{\textsc{LBM-Exact}\xspace}
\newcommand{\Md}{M_d}
\newcommand{\Ms}{M_s}
\newcommand{\Vd}{\mathcal{V}_d}
\newcommand{\Vs}{\mathcal{V}_s}

\newcommand{\dsafe}{\Delta_{\text{safe}}}

\title{\includegraphics[width=1\linewidth]{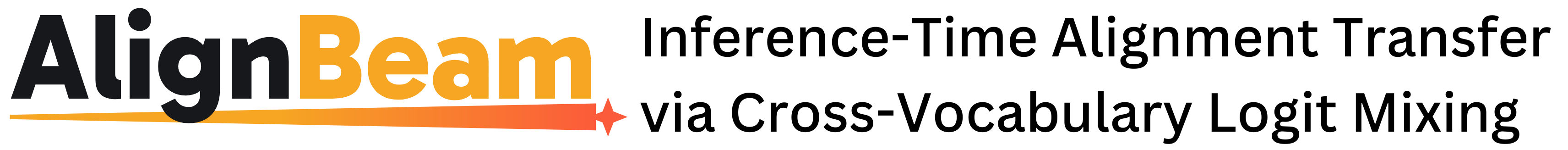}}

\author{%
  Chirag Chawla\thanks{Work done while at Lexsi Labs.},
  Pratinav Seth,
  Vinay Kumar Sankarapu \\
  \affiliation{Lexsi Labs} \\
  \texttt{pratinav.seth@lexsi.ai}
}

\setabstract{%
Domain fine-tuning degrades the safety of large language models:
fine-tuned specialists readily comply with harmful prompts framed
in domain language. Existing inference-time defenses that mix logits
from a safe \emph{anchor} model require both models to share a
vocabulary, which rules them out for the cross-family specialists
where safety is most degraded. We present \alignbeam, a
training-free method that lifts this restriction by translating
anchor logits into the target model's vocabulary token-by-token at
each decoding step; a small LLM judge then selects the safest among
$K$ candidate continuations. No weights are changed, and the
safety-utility trade-off can be tuned at deployment without
retraining. Across both cross-vocabulary and same-vocabulary
evaluation pairs, \alignbeam substantially raises refusal on
adversarial benchmarks while keeping task accuracy and inference
overhead within practical bounds. The results show that safety
alignment can be transferred between model families at inference
time, without touching either model's weights.
}

\setkeywords{safety alignment, inference-time defense, cross-vocabulary decoding,
  logit mixing, domain fine-tuning, alignment transfer, LLM safety, beam search}

\runningtitle{AlignBeam: Inference-Time Alignment Transfer}

\begin{document}
\maketitle

\section{Introduction}
\label{sec:intro}

Domain fine-tuning degrades safety alignment. A medical specialist
trained on patient notes will comply with pharmacology abuse questions
framed as clinical queries; a finance specialist will help plan fraud
once the harm is framed as a portfolio problem. System prompts offer
no reliable fix: a system-prompted Llama-3.1-8B achieves 14.3\%
refusal, \emph{below} the unprompted draft's 16.9\%
(\S\ref{sec:results}). 

Post-generation classifiers can flag harmful
outputs but cannot steer generation away from them, and RLHF/DPO
retraining is undone by as few as 100 benign fine-tuning steps
\cite{qi2023finetuning}. A more direct fix is to mix logits from a
small, well-aligned anchor model during generation, but existing
methods \cite{xu2024safedecoding,liu2024proxy,li2023contrastive}
require both models to share a vocabulary. This rules them out for
the cross-family specialists where safety is most degraded.

Safety alignment concentrates in the first few output tokens and is
fragile to domain shift \cite{qi2025depth}, which is why
mixing only a short prefix is enough. \alignbeam is inspired by \textsc{SafeDecoding}
\cite{xu2024safedecoding} and extends it to the cross-vocabulary
setting: when the two models share a tokeniser, $\Vd\!=\!\Vs$,
\lbm reduces to a construction analogous to \textsc{SafeDecoding},
with the difference that the support is selected via beam search
rather than token-level intersection. A
full comparison with LlamaGuard \cite{inan2023llamaguard},
\textsc{RAIN} \cite{li2024rain}, hard-prefix priming, and recent
decoding-time methods
\cite{fei2025nudging,huang2025deal,lyu2025backtracking} is in
Appendix~\ref{app:related-extended}.

We introduce \alignbeam\footnote{Code will be released at \url{https://github.com/Lexsi-Labs/alignbeam}.}, a training-free method that works as follows. At each decoding step, the anchor model's top predicted tokens are decoded to text and re-encoded in the draft model's tokeniser, projecting the anchor's next-token distribution into draft vocabulary space. A mixing weight $\alpha$ blends this translated distribution with the draft model's own, steering early tokens toward safe completions without any shared vocabulary requirement. Three candidate continuations are scored by a small LLM judge, and the safest is returned. Mixing depth $N$, weight $\alpha$, and beam count $K$ are all tunable at deployment without retraining (\S\ref{sec:method}). 

We evaluate across five model pairs covering three SFT-eroded domain specialists, one unaligned base model, and a same-vocabulary control, and find consistent, large gains on adversarial benchmarks with minimal accuracy cost (\S\ref{sec:results}).

\paragraph{Contributions.}
\begin{enumerate}[leftmargin=*,itemsep=1pt,topsep=2pt]
  \item \textbf{Cross-vocabulary logit bridge mixing (\lbm).}
    A text-bridge construction (decode anchor token, re-encode under
    the draft tokeniser) that enables probability mixing across
    heterogeneous tokenisers without shared token IDs, in three
    multi-token fallback variants (\S\ref{sec:method}).
  \item \textbf{Three-phase beam pipeline with deployment-time knobs.}
    Mixing depth $N$ trades safety against speed and weight $\alpha$
    trades safety against utility, both tunable without retraining
    (\S\ref{sec:method}).
  \item \textbf{Early-token safety, confirmed.}
    Beam priming with three mixed steps ($N\!=\!0\!\to\!3$) captures
    $+$61.9\,pp of the HB$+$AB safety gain, after which returns
    diminish sharply (\S\ref{sec:results}).
  \item \textbf{Five-pair empirical validation.}
    Medical, financial, math, and general drafts in base/instruct and
    same/cross-vocabulary regimes, with additional 70B-scale,
    LoRA-degradation, and anchor-size studies (\S\ref{sec:setup}).
\end{enumerate}

\section{Related Work}
\label{sec:related}

\noindent\textbf{Alignment erosion under fine-tuning.}
As few as 100 benign fine-tuning samples can substantially reduce
refusal \cite{qi2023finetuning}, and authority and domain framing
provide potent soft jailbreaks against aligned models
\cite{yang2023shadow}. Domain-specialist training is far
more extensive than these small-scale demonstrations, and our
five-pair evaluation correspondingly observes baseline refusal rates
ranging from near-zero to moderate across domains.

\noindent\textbf{Inference-time safety steering.}
\textsc{SafeDecoding} \cite{xu2024safedecoding} is our closest prior
work: it blends draft and safety-model probabilities at their shared
vocabulary intersection and so does not apply to cross-family pairs.
\alignbeam relaxes this assumption via the text bridge and additionally
targets base models without chat-template conditioning; when
$\Vd\!=\!\Vs$ it reduces to a construction analogous to
\textsc{SafeDecoding} (differing in support selection: beam search
vs.\ token-level intersection). \textsc{Nudging} \cite{fei2025nudging}
performs a binary swap to a smaller aligned guide on high-entropy
steps, where we instead use a continuous $\alpha$-weighted blend with
beam search and LLM-judge selection. \textsc{Proxy Tuning}
\cite{liu2024proxy} and top-$k$ contrastive decoding
\cite{li2023contrastive} apply the same logit arithmetic but presume a
shared vocabulary; on the same-vocabulary Qwen3-8B-Base pair \alignbeam
outperforms both, most sharply on contextual and Sorry-Bench
categories (Appendix~\ref{app:eq1-methods}). \textsc{RAIN}
\cite{li2024rain} self-aligns via rewindable generation but needs a
safety prior in the draft to produce the signals that drive rewinding,
which is exactly what domain fine-tuning removes.

\noindent\textbf{Early-token safety concentration.}
Safety alignment concentrates in the first few token positions and is
therefore brittle to prompt and decoding shifts \cite{qi2025depth}. Our depth ablation corroborates this directly and
motivates the small mixed depths ($N\!\in\!\{3,6\}$) used throughout.
A fuller treatment, covering post-hoc guard classifiers
\cite{inan2023llamaguard} and recent decoding-time methods
\cite{huang2025deal,lyu2025backtracking}, is in
Appendix~\ref{app:related-extended}.

\begin{figure}[pt]
\centering
\includegraphics[width=0.99\linewidth]{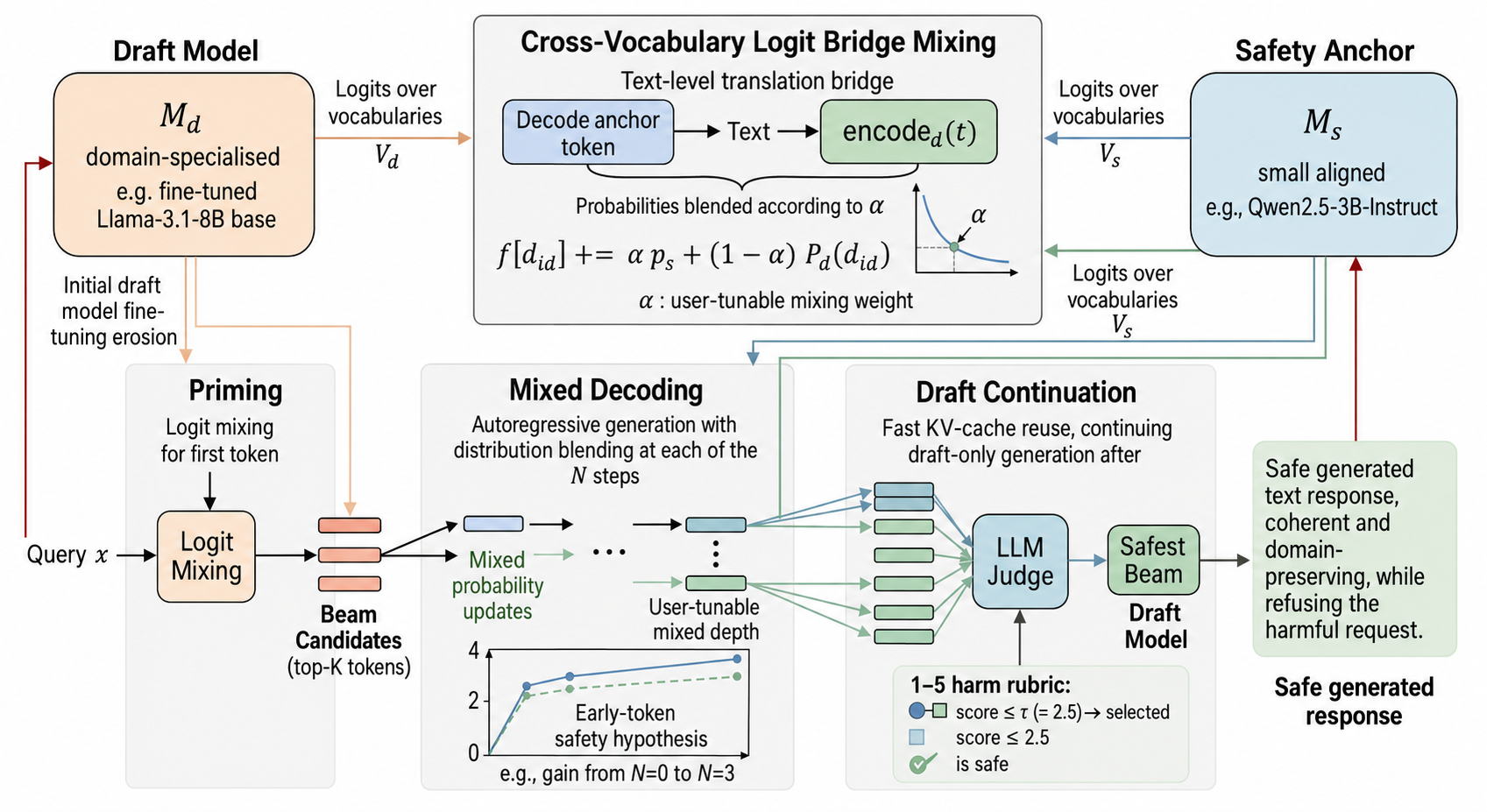}
\vspace{6.3pt}
\caption{Anchor logits govern the safety-critical
  opening positions; the draft model takes over for fluency and
  domain accuracy. \textsc{Phase~1 (Priming):} \lbm at step~0
  yields $K$ beam roots.
  \textsc{Phase~2 (Mixed Decoding):} each beam is extended greedily
  for $N{-}1$ steps via \lbm. \textsc{Phase~3 (Draft Continuation):}
  an LLM judge picks the safest of $K$ beams; $\Md$ alone continues
  it for domain quality (KV-cache reuse).}
\label{fig:overview}
\end{figure}

\section{Method}
\label{sec:method}
Let $\Md$ be a domain-specialist draft model with vocabulary $\Vd$
and $\Ms$ a small aligned safety anchor with vocabulary $\Vs$;
in general $\Vd \neq \Vs$. Given a query $x$, we seek a response that
is both \emph{safe} and \emph{useful}. Neither model is modified;
\alignbeam runs at inference time
(Figure~\ref{fig:overview}).

\noindent\textsc{Cross-vocabulary logit bridge mixing (\lbm).}
Because $\Md$ and $\Ms$ use different vocabularies, token IDs cannot
be mixed directly. \lbm bridges them by translating the anchor's
top-$B$ predicted tokens (we use $B\!=\!50$) into draft vocabulary
space. 

Let $P_s$, $P_d$ denote the next-token distributions and
$\mathcal{T}_B \subset \Vs$ the anchor's top-$B$ tokens. For each
$s_{\mathrm{id}} \in \mathcal{T}_B$ with probability
$p_s = P_s(s_{\mathrm{id}})$, we decode it to a string and
re-encode under the draft tokeniser:
$t = \mathrm{decode}_s(s_{\mathrm{id}})$,
$d_{\mathrm{ids}} = \mathrm{encode}_d(t)$. Whenever the result
is a single draft token ($|d_{\mathrm{ids}}|\!=\!1$), we accumulate
it into a buffer $f$ via
\begin{equation}
  f[d_{\mathrm{id}}] \mathrel{+}=
    \alpha\, p_s + (1-\alpha)\, P_d(d_{\mathrm{id}}),
  \label{eq:lbm-mix}
\end{equation}
where $\alpha \in [0,1]$ is a deployment-time safety-utility knob:
higher $\alpha$ steers more aggressively toward the anchor's
distribution; at $\alpha\!=\!0$ the anchor contributes no logit
signal, though the text-bridge mapping still executes.
When the string decodes to multiple draft tokens, a
variant-specific fallback applies: \lbmdrop skips them entirely,
\lbmfirst keeps only the first sub-token, and \lbmexact discards.

Renormalising $f$ gives the blended distribution $\tilde{f}$. On
same-vocabulary pairs all three variants agree ($>$98\% single-token
match); on cross-family pairs roughly 12\% of anchor tokens hit the
fallback. When $\Vd\!=\!\Vs$, \lbm is analogous to
\textsc{SafeDecoding} \cite{xu2024safedecoding}; the construction
differs in that \lbm selects support via beam search over
continuations rather than by token-level top-$k$ intersection.

We use \lbmdrop as the default: it carries the full anchor
distribution at each mixed position, suppressing harmful token mass
and producing standard refusal phrases. \lbmfirst injects only
the first sub-token per step, steering the output toward safety
without triggering the string-match refusal list (\emph{stealth
safety}; Appendix~\ref{app:e2-strategies}).

\begin{table}[htbp]
\centering\footnotesize
\caption{The three \lbm variants and their multi-token
  fallback rules. \lbmdrop is the default, \lbmfirst gives
  \emph{stealth safety} and leads on DeepSeek-Math-7B-Instruct, and \lbmexact is a
  conservative ablation. On same-vocabulary pairs all three coincide
  ($>$98\% single-token match).}
\label{tab:variants}
\vspace{6.3pt}
\begin{tabular}{@{}lll@{}}
\toprule
\textbf{Variant} & \textbf{Multi-token fallback} & \textbf{Use case} \\
\midrule
\lbmdrop  & Drop; renormalise     & All drafts (default) \\
\lbmfirst & Use first sub-token   & Stealth safety / instruct \\
\lbmexact & Keep only 1:1 matches & Conservative ablation \\
\bottomrule
\end{tabular}%
\end{table}

\noindent\textsc{Three-phase beam pipeline.}
\begin{itemize}
    \item \emph{Phase~1} runs both models on query $x$ (chat template for instruct models; \texttt{Q:\{x\}\textbackslash nA:} for base) and picks the top-$K$ tokens of $\tilde{f}$ as starting beams.
    \item \emph{Phase~2} extends each beam greedily for $N{-}1$ further steps under \lbm, requiring $O(K\!\cdot\!N)$ paired forward passes in total. 
    \item \emph{Phase~3} completes all $K$ beams using $\Md$ alone with KV-cache reuse; a small LLM judge $J$ scores each beam on a 1--5 harm rubric and returns the safest one ($\mathrm{score}\!\le\!\tau\!=\!2.5$).
\end{itemize}
 Phase~3 is the main computational cost, so total overhead scales with $K$, not $N$.
We offer two settings: $K\!=\!1$ (no judge, lowest overhead) and
$K\!=\!3$ (default), both within the latency range of existing
inference-time defenses (Appendix~\ref{app:latency}).
\vspace{10pt}
\begin{algorithm}[htbp]
\footnotesize
\caption{\alignbeam Generation}
\label{alg:alignbeam}
\begin{algorithmic}[1]
\Require Query $x$; draft $\Md$; anchor $\Ms$;
         depth $N$; beams $K$; weight $\alpha$
\Ensure Response $y^*$
\Statex \emph{Phase 1: Priming}
\State $x_d \gets \mathrm{fmt}_d(x)$;\enspace $x_s \gets \mathrm{fmt}_s(x)$
\State $\tilde{f} \gets \mathrm{LBM}(\Md(x_d),\,\Ms(x_s),\,\alpha)$
\State $\{b_k\}_{k=1}^{K} \gets \text{top-}K\text{ tokens of }\tilde{f}$
\Statex \emph{Phase 2: Mixed Decoding}
\For{step $= 1, \ldots, N{-}1$; each beam $b_k$}
  \State $\tilde{f}^{(k)} \gets \mathrm{LBM}(\Md(x_d{\|}b_k),\,\Ms(x_s{\|}b_k),\,\alpha)$
  \State $v^* \gets \arg\max_v \tilde{f}^{(k)}(v)$;\;\;extend $b_k$
\EndFor
\Statex \emph{Phase 3: Draft Continuation}
\For{each beam $b_k$}
  \State Complete $b_k$ with $\Md$ alone (KV-cache reuse)
\EndFor
\State $b^* \gets \arg\min_{k:\,J(b_k)\le\tau} J(b_k)$
\State \Return $y^* \gets \mathrm{text}(b^*)$
\end{algorithmic}
\end{algorithm}

\section{Experiments}
\label{sec:setup}

\textbf{Model pairs.}
We evaluate five drafts ranging from 7B to 8B parameters: three
domain-fine-tuned specialists where SFT degraded safety
(\textsc{MedLlama-3-8B, Finance-Llama3-8B, DeepSeek-Math-7B-Instruct}), one unaligned base model used as
reference control (\textsc{Llama-3.1-8B}), and one same-vocabulary
control (\textsc{Qwen3-8B-Base}, which shares a tokeniser with the
anchor).
DeepSeek-Math-7B-Instruct uses $N\!=\!20$ rather than the default
$N\!=\!6$: at $N\!=\!6$ it produces a refusal prefix and then
continues with the harmful answer, so a longer mixing window is
needed (Appendix~\ref{app:e2-strategies}). The safety anchor is
Qwen2.5-3B-Instruct, fixed across all pairs
(Table~\ref{tab:model-pairs}).

\begin{table}[pt]
\centering\footnotesize
\setlength{\tabcolsep}{4pt}
\caption{The five draft/anchor pairs in our evaluation.
  The safety anchor is Qwen2.5-3B-Instruct for every pair.
  Defaults: $\alpha\!=\!0.5$, $K\!=\!3$ (safety) / $K\!=\!1$ (utility),
  seed~42. DeepSeek-Math-7B-Instruct uses $N\!=\!20$ to suppress a refuse-then-comply pattern
  seen at $N\!=\!6$; Qwen3-8B-Base is the only same-vocabulary pair.}
\label{tab:model-pairs}
\vspace{6.3pt}
\begin{tabular}{@{}lllcr@{}}
\toprule
\textbf{Draft model} & \textbf{Dom.}
  & \textbf{Type} & \textbf{Voc.} & \textbf{$N$} \\
\midrule
DeepSeek-Math-7B-Instruct & Math & Inst. & Cross          & 20 \\
Llama-3.1-8B        & Gen. & Base  & Cross          & 6  \\
MedLlama-3-8B       & Med. & Base  & Cross          & 6  \\
Finance-Llama3-8B   & Fin. & Base  & Cross          & 6  \\
Qwen3-8B-Base       & Gen. & Base  & \textbf{Same}  & 6  \\
\bottomrule
\end{tabular}
\end{table}

\textbf{Datasets.}
We use five complementary evaluation axes (sample counts in
parentheses). \textsc{Safety:}
HarmBench-Standard (167) \cite{mazeika2024harmbench},
HarmBench-Contextual (62), AdvBench (520) \cite{zou2023adversarial},
Sorry-Bench (440) \cite{xie2024sorrybench},
WildJailbreak-Eval-Harmful (1{,}105--1{,}315) \cite{ding2024wildjailbreak}.
\textsc{Calibration (benign):}
XSTest (450) \cite{rottger2023xstest},
OR-Bench-Hard (855) \cite{cui2024orbench}, JBB-Benign (100) \cite{chao2024jailbreakbench}.
\textsc{Utility:} GSM8K \cite{cobbe2021gsm8k} for DeepSeek-Math-7B-Instruct,
MedQA \cite{jin2021medqa} for MedLlama-3-8B.
HarmBench-Contextual and WildJailbreak probe domain-framed and
multi-sentence attacks respectively; calibration sets measure benign
over-refusal. Supplementary harmful sets
(JailbreakBench-Harmful~\cite{chao2024jailbreakbench})
are reported only in the appendix.

\textbf{Hyperparameters and judging.}
All runs use: $N\!=\!6$ (DeepSeek-Math: $20$), $K\!=\!3$ for safety
benchmarks and $K\!=\!1$ for utility, $\alpha\!=\!0.5$, judge
threshold $\tau\!=\!2.5$, temperature $T\!=\!0.7$, bridge width
$B\!=\!50$, repetition penalty $1.15$, seed $42$. Each benchmark
is scored by its own official judge (Appendix~\ref{app:judges}).
All numbers reported in the paper are \emph{string-match refusal
rates}, a conservative primary metric whose rationale is in
Appendix~\ref{app:string-vs-acc}. The LLM judge in Phase~3 is used
\emph{only} to pick between beams during generation; it never
scores the results reported here, avoiding any self-evaluation
circularity (full per-benchmark setup in
Appendix~\ref{app:judges}).

\textbf{Why string-match refusal is conservative.}
Base models sometimes emit incoherent or repetitive text on
adversarial prompts, which official judges correctly score as
non-harmful; this inflates the apparent baseline (on Llama-3.1-8B, string-match
refusal is 16.9\% against 49.1\% judge accuracy, a $\approx$32\,pp gap
from degenerate outputs counted ``safe''). \alignbeam's outputs are
non-degenerate by construction, since the judge filters incoherent
beams, so the gap between Base and \alignbeam under string-match is a
lower bound on the true safety gain. We report string-match refusal as
the primary metric and benchmark-judge accuracy as secondary; the full
analysis is in Appendix~\ref{app:string-vs-acc}.

\section{Results}
\label{sec:results}

\subsection{Headline Results}
Across five pairs, \alignbeam raises string-match refusal on
\textbf{AdvBench} from 38.1\% to \textbf{91.5\%} ($+$53.4\,pp
macro avg.) and on \textbf{HarmBench-Std} from 25.9\% to
\textbf{76.4\%} ($+$50.5\,pp macro avg.)
(Table~\ref{tab:headline}; both share the HarmBench-Mistral judge).
Four pairs reach \textbf{91--96\% on AdvBench} and \textbf{78--81\%
on HarmBench-Std}; DeepSeek-Math-7B-Instruct gains $+$64.9\,pp, bringing the
macro average to 76.4\% (median 79.6\%). \textsc{Benign over-refusal} (refusal on non-adversarial prompts;
lower is better) rises across all pairs as the anchor's tendency
to refuse carries over (Llama-3.1-8B: OR-H 11.0\%$\to$22.3\%, XSTest
5.6\%$\to$40.0\%; per-pair details in
Appendix~\ref{app:e3-breakdown}). \textsc{Task utility} falls by
$-$0.4\,pp on both GSM8K and MedQA (Appendix~\ref{app:utility}).

\begin{table}[pt]
\centering
\footnotesize
\caption{\textbf{\alignbeam raises string-match refusal by $+$35--74\,pp
  on cross-vocabulary pairs and holds benign over-refusal within the LlamaGuard band.}
  Ref\% = string-match refusal (bold, primary); Acc\% = benchmark-judge accuracy (secondary);
  \underline{underlined} = improvement over the unmodified draft.
  Config: \lbmdrop,
  $\alpha\!=\!0.5$, $N\!=\!6$ (DeepSeek-Math: $N\!=\!20$), $K\!=\!3$, seed~42.
  Full Base/\alignbeam grid in Appendix~\ref{app:main-safety-full}.}
\label{tab:headline}
\vspace{6.3pt}
\begin{tabular}{@{}l l rrrrr@{}}
\toprule
 & & \multicolumn{5}{c}{\textbf{\alignbeam on harmful benchmark} ($\uparrow$)} \\
\cmidrule(lr){3-7}
\textbf{Draft model (domain)} &
  & \textbf{HB-Std} & \textbf{HB-Ctx} & \textbf{AdvB}
  & \textbf{Sorry} & \textbf{WildJB} \\
\midrule
\multicolumn{7}{@{}l}{\textit{Domain-fine-tuned specialists (safety eroded by SFT; \alignbeam restores it)}}\\[1pt]
\multirow{2}{*}{MedLlama-3-8B (medical)}
  & {\scriptsize Ref\%} & \textbf{78.4}\,{\scriptsize\underline{$+$61.6}} & \textbf{69.4}\,{\scriptsize\underline{$+$59.7}} & \textbf{91.7}\,{\scriptsize\underline{$+$51.5}} & \textbf{63.2}\,{\scriptsize\underline{$+$45.2}} & \textbf{16.2}\,{\scriptsize\underline{$+$2.1}} \\
  & {\scriptsize Acc\%} & {\scriptsize 88.6} & {\scriptsize 87.1} & {\scriptsize 99.0} & {\scriptsize 72.3} & {\scriptsize 27.1} \\[1pt]
\multirow{2}{*}{Finance-Llama3-8B (finance)}
  & {\scriptsize Ref\%} & \textbf{79.6}\,{\scriptsize\underline{$+$56.2}} & \textbf{80.6}\,{\scriptsize\underline{$+$50.0}} & \textbf{94.8}\,{\scriptsize\underline{$+$65.2}} & \textbf{65.7}\,{\scriptsize\underline{$+$10.5}} & \textbf{33.2}\,{\scriptsize($-$9.5)} \\
  & {\scriptsize Acc\%} & {\scriptsize 91.0} & {\scriptsize 90.3} & {\scriptsize 98.1} & {\scriptsize 77.3} & {\scriptsize 41.4} \\[1pt]
\multirow{2}{*}{DeepSeek-Math-7B-Instruct (math)}
  & {\scriptsize Ref\%} & \textbf{63.5}\,{\scriptsize\underline{$+$50.9}} & \textbf{61.3}\,{\scriptsize\underline{$+$51.6}} & \textbf{80.6}\,{\scriptsize\underline{$+$69.4}} & \textbf{45.0}\,{\scriptsize\underline{$+$31.8}} & \textbf{15.0}\,{\scriptsize\underline{$+$3.5}} \\
  & {\scriptsize Acc\%} & {\scriptsize 85.0} & {\scriptsize 79.0} & {\scriptsize 91.3} & {\scriptsize 65.7} & {\scriptsize 25.6} \\
\midrule
\multicolumn{7}{@{}l}{\textit{Pre-trained / unaligned drafts (reference controls)}}\\[1pt]
\multirow{2}{*}{Llama-3.1-8B (general)}
  & {\scriptsize Ref\%} & \textbf{79.6}\,{\scriptsize\underline{$+$64.6}} & \textbf{79.0}\,{\scriptsize\underline{$+$70.9}} & \textbf{94.4}\,{\scriptsize\underline{$+$76.9}} & \textbf{61.4}\,{\scriptsize\underline{$+$50.5}} & \textbf{19.7}\,{\scriptsize\underline{$+$6.4}} \\
  & {\scriptsize Acc\%} & {\scriptsize 91.6} & {\scriptsize 80.7} & {\scriptsize 98.3} & {\scriptsize 74.3} & {\scriptsize 30.1} \\[1pt]
\multirow{2}{*}{Qwen3-8B-Base (same-vocab ctrl)}
  & {\scriptsize Ref\%} & \textbf{80.8}\,{\scriptsize\underline{$+$19.1}} & \textbf{71.0}\,{\scriptsize\underline{$+$33.9}} & \textbf{95.8}\,{\scriptsize\underline{$+$3.9}} & \textbf{63.4}\,{\scriptsize\underline{$+$23.9}} & \textbf{20.9}\,{\scriptsize\underline{$+$4.8}} \\
  & {\scriptsize Acc\%} & {\scriptsize 89.8} & {\scriptsize 90.3} & {\scriptsize 99.4} & {\scriptsize 79.8} & {\scriptsize 32.6} \\
\midrule
\multirow{2}{*}{\textbf{Macro avg.}}
  & {\scriptsize Ref\%} & \textbf{76.4}\,{\scriptsize\underline{$+$50.5}} & \textbf{72.3}\,{\scriptsize\underline{$+$53.2}} & \textbf{91.5}\,{\scriptsize\underline{$+$53.4}} & \textbf{59.7}\,{\scriptsize\underline{$+$32.4}} & \textbf{21.0}\,{\scriptsize\underline{$+$1.5}} \\
  & {\scriptsize Acc\%} & {\scriptsize 89.2} & {\scriptsize 85.5} & {\scriptsize 97.2} & {\scriptsize 73.9} & {\scriptsize 31.4} \\
\bottomrule
\end{tabular}
\end{table}

\begin{figure}[pt]
\centering
\includegraphics[width=0.81\linewidth]{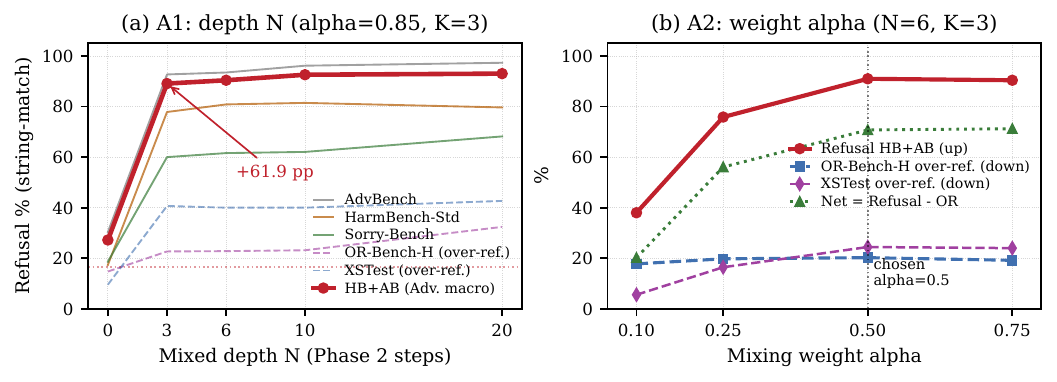}
\vspace{6.3pt}
\caption{Three mixed steps capture $>$92\% of the
  safety gain; benign over-refusal is insensitive to the mixing
  weight $\alpha$.
  \textbf{(a) Depth sweep} ($\alpha\!=\!0.85$, $K\!=\!3$):
  HB$+$AB elbows sharply at $N\!=\!3$ ($+$61.9\,pp over $N\!=\!0$);
  bold red $=$ HB$+$AB; thin solid $=$ per-benchmark; dashed $=$
  benign over-refusal (lower is better).
  \textbf{(b) Weight ($\alpha$) sweep} ($N\!=\!6$, $K\!=\!3$): refusal
  saturates by $\alpha\!=\!0.50$ while OR-Bench-Hard over-refusal
  stays flat, showing the ceiling is set by anchor calibration, not
  $\alpha$.
  Net $=$ Refusal $-$ over-refusal.}
\label{fig:ablations}
\end{figure}

\subsection{Early-Token Safety and the Depth/Weight Trade-off}
The depth ablation (Llama-3.1-8B, $\alpha\!=\!0.85$, $K\!=\!3$) confirms
that safety is concentrated in early tokens \cite{qi2025depth}:
beam priming alone ($N\!=\!0$) lifts AdvBench refusal from 17.5\%
to 30.4\%, the first three mixed steps add \textbf{$+$62.3\,pp}
(reaching 92.7\%), and going deeper than $N\!=\!6$ adds only
$\approx$3.9\,pp more on HB$+$AB (Figure~\ref{fig:ablations}(a)). The weight
ablation ($N\!=\!6$, $K\!=\!3$) shows refusal saturating by
$\alpha\!=\!0.50$ (AdvBench 94.4\%, HarmBench-Std 80.2\%) while
OR-Bench-Hard over-refusal stays flat at $\approx$20\% across all
$\alpha$ (\S\ref{sec:limits}). The $\alpha$ knob
thus controls where on the refusal/over-refusal curve to operate
(Figure~\ref{fig:ablations}(b)).
We use $\alpha\!=\!0.50$, $N\!=\!6$ as defaults; full sweeps are in
Appendices~\ref{app:a1-full}--\ref{app:a2-full}.

\subsection{Beam Count $K$}
$K$ is set once at deployment. $K\!=\!1$ (no judge, lowest cost)
captures $\sim$80\% of the safety gain at roughly $2\times$
overhead, comparable to hard-prefix priming. $K\!=\!3$ (default,
with LLM judge) recovers the remaining gain at $\sim$5$\times$
overhead. Per-pair latency breakdowns are in
Appendix~\ref{app:latency}.

\subsection{Anchor Alignment vs.\ Beam Search}
\begin{figure}[pt]
\centering
\includegraphics[width=0.51\linewidth]{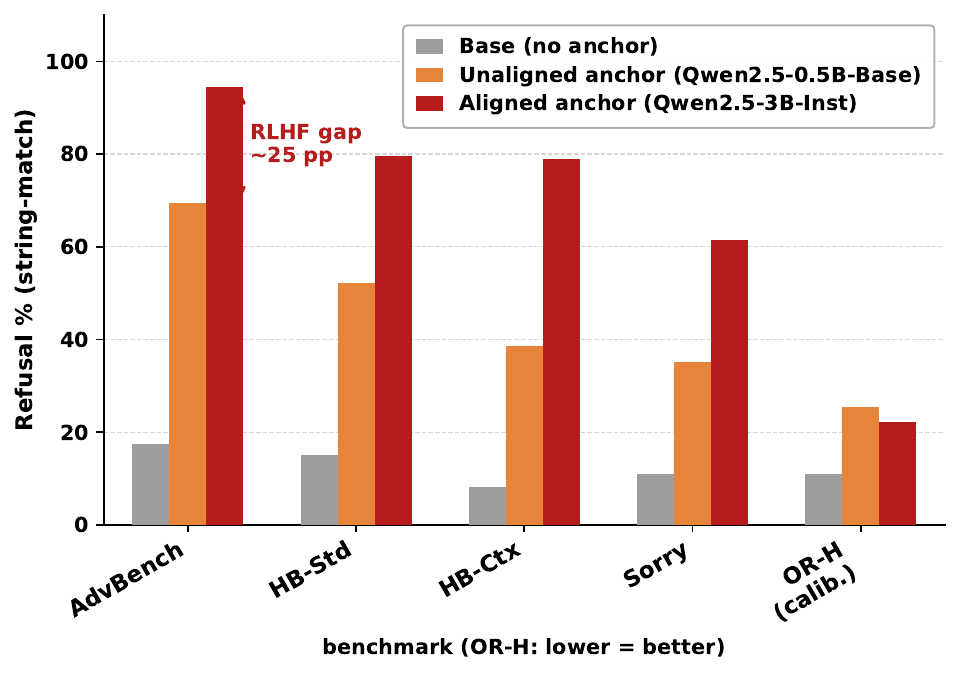}
\vspace{6.3pt}
\caption{Anchor \emph{alignment} (RLHF) adds
  $\sim$+25\,pp over beam search alone at no extra cost.
  Anchor-alignment ablation on Llama-3.1-8B ($N\!=\!6$, $K\!=\!3$,
  $\alpha\!=\!0.5$): beam mechanism alone (unaligned anchor)
  gives $\sim$25--52\,pp on harmful sets; the RLHF-aligned
  anchor of the same size adds another $\sim$+25\,pp.
  Last group (OR-H) is benign over-refusal: lower is better.}
\label{fig:anchor-rlhf}
\end{figure}

On Llama-3.1-8B, beam search and anchor alignment each contribute roughly half
the safety gain. Replacing the aligned anchor with its unaligned base
counterpart leaves $\sim$25--52\,pp refusal from beam search alone;
swapping back to the aligned version adds another $+$25\,pp for free
(Figure~\ref{fig:anchor-rlhf}). A separate anchor-size sweep
(Qwen2.5-7B-Base draft) shows that alignment quality matters more
than model size: a 0.5B aligned anchor reaches 80.2\% HB-Std judge
accuracy versus 98.2\% for the 7B anchor, with the gap driven by
calibration rather than parameter count (Appendix~\ref{app:eq2}).

\begin{table}[pt]
\centering
\footnotesize
\caption{\alignbeam matches LlamaGuard on adversarial
  safety (78--94\% refusal) while keeping benign over-refusal below
  27\% vs.\ LlamaGuard's 34--39\%. String-match refusal\%; harmful:
  $\uparrow$; $^{\dagger}$benign / calibration (over-refusal): $\downarrow$.
  All rows: same Llama-3.1-8B draft. Hard-prefix at $K\!=\!1$;
  \alignbeam at $K\!=\!3$.}
\label{tab:e3-baselines}
\vspace{6.3pt}
\begin{tabular}{@{}l rrrr c rr@{}}
\toprule
 & \multicolumn{4}{c}{\textbf{Harmful} ($\uparrow$)}
 & & \multicolumn{2}{c}{\textbf{Benign}$^{\dagger}$ ($\downarrow$)} \\
\cmidrule(lr){2-5}\cmidrule(lr){7-8}
\textbf{Method}
  & \textbf{HB-Std} & \textbf{HB-Ctx} & \textbf{AdvB} & \textbf{Sorry}
  & & \textbf{OR-H} & \textbf{JBB-B} \\
\midrule
Base draft                 & 15.0 &  8.1 & 17.5 & 10.9 & & 11.0 & 10.0 \\
Safety prompt              & 10.2 &  8.1 & 15.6 & 15.2 & & 14.7 & 10.0 \\
Hard prefix                & 82.0 & 85.5 & 97.7 & 63.0 & & 23.7 & 29.0 \\
LlamaGuard prompt-filter   & 98.2 & 95.2 & 99.4 & 85.5 & & 34.5 & 38.0 \\
LlamaGuard response-filter & 98.2 & \textbf{100.0} & 99.4 & 83.4 & & 34.6 & 39.0 \\
\textbf{\alignbeam} (ours) & \textbf{79.6} & 79.0 & \textbf{94.4} & \textbf{61.4} & & \textbf{22.3} & \textbf{26.0} \\
\bottomrule
\end{tabular}%
\end{table}

\subsection{Comparison to Baselines}
LlamaGuard filters reach near-ceiling on adversarial benchmarks but
push benign over-refusal to $\approx$34.5\% on OR-Bench-Hard and 38--39\%
on JBB-Benign. \alignbeam achieves 79.6\% HarmBench-Std and 94.4\%
AdvBench refusal while keeping OR-Bench-Hard at 22.3\% and JBB-Benign
at 26.0\%, a substantially better safety/over-refusal balance
(Table~\ref{tab:e3-baselines}; Appendix~\ref{app:llamaguard-baselines}).
The safety-prompt baseline is actively counterproductive: 14.3\%
refusal, \emph{below} the unprompted draft's 16.9\%. On the
same-vocabulary Qwen3-8B-Base pair, \alignbeam beats both
\textsc{Proxy Tuning} \cite{liu2024proxy} and top-$k$ contrastive
decoding \cite{li2023contrastive} on five of six adversarial
benchmarks, with the widest gap on contextual attacks
(Appendix~\ref{app:eq1-methods}); results hold at 70B scale
(Llama-3.1-70B, Appendix~\ref{app:e4}).

\subsection{Robustness}
Results are stable across seeds \{42,\,2,\,3\} ($\pm$0.7\,pp
HB$+$AB; Appendix~\ref{app:seed}) and across different Phase~1 prompt
formats (Appendix~\ref{app:a7}). Dropping the LLM judge altogether
($K\!=\!1$, Appendix~\ref{app:nojudge}) leaves adversarial refusal
unchanged or slightly higher; the only cost is $+$6.4\,pp XSTest
over-refusal, confirming that logit mixing provides the safety
signal and the judge's role is primarily to reduce over-refusal on
benign inputs. Full ablations and per-pair breakdowns are in the appendix.

\subsection{Task Utility}
Under \alignbeam's $K\!=\!1$ utility mode, task accuracy drops by
$-$0.4\,pp on both domain pairs: GSM8K (DeepSeek-Math-7B-Instruct)
77.0\%\,$\to$\,76.6\% and MedQA (MedLlama-3-8B)
13.1\%\,$\to$\,12.7\%. Full per-strategy MedQA similarity metrics
(ROUGE-1/2, BERTScore-F1, SentSim; all within $\pm$1.5\,pp) are
in Appendix~\ref{app:medqa-semantic}.

\section{Conclusion}
\label{sec:conclusion}

Safety alignment is not an intrinsic property of a model's weights:
it can be loaned at inference time from a separately maintained anchor,
decoupling safe deployment from the fine-tuning pipeline.
\alignbeam instantiates this principle via a cross-vocabulary text
bridge, recovering near-complete AdvBench refusal on cross-vocabulary
specialist pairs and substantially lifting count-weighted HB$+$AB
refusal, while holding task accuracy within a fraction of a percentage
point and fitting within the latency band of existing defenses.
Immediate extensions include multi-anchor ensembles, adaptive-prefix
decoding, and a benign-prompt calibration head to reduce over-refusal.

\section*{Limitations}
\label{sec:limits}

\noindent\textsc{Compute overhead.}
Inference cost is comparable to other inference-time defenses
(LlamaGuard, Proxy Tuning, hard-prefix, RAIN); $K$ and $N$ are
deployment-adjustable without retraining (Appendix~\ref{app:latency}).

\noindent\textsc{Calibration ceiling.}
Both safety limits stem from the anchor, not the mixing parameters:
the anchor's over-refusal on ambiguous benign prompts is inherited
(OR-Bench-H $\approx$20\%, XSTest $\approx$40\% on Llama-3.1-8B), and the
mixed prefix covers only the first $N$ tokens, so attacks that
embed harm past the prefix are only partially blocked
(WildJailbreak-H: $+$2--6\,pp vs.\ $+$51--77\,pp on AdvBench). A
better-calibrated anchor, longer prefix, or per-prompt $\alpha$
schedule could close the gap.

\noindent\textsc{Evaluation scope.}
All prompts are single-turn and $\leq$200 tokens; multi-turn
jailbreaks and adaptive attacks are out of scope. Experiments use
a single Qwen2.5-3B-Instruct anchor; the alternative-anchor study
(Appendix~\ref{app:a8a}) shows results depend on anchor refusal style.
Multi-anchor ensembles, multilingual evaluation, and human judging
are natural next steps. Truncation and ``sorry-but-continues''
artefacts that affect verbose drafts and understate true safety are
discussed in Appendix~\ref{app:ext-limits}.

\section*{Ethics Statement}

\alignbeam is designed to reduce harmful outputs from domain-fine-tuned
language models. Because it raises refusal rates rather than
suppressing them, direct misuse is limited, though the method relies
on a well-aligned anchor: a misaligned one could itself steer
generation toward harm. The main benefit is
allowing organisations to deploy specialised models without
compromising their safety. Inference overhead can be adjusted via
$K$ and $N$ to fit a deployment budget (Appendix~\ref{app:latency});
all experiments ran on a single RTX~6000 Pro Blackwell (96~GB, bfloat16).
All evaluation is automated and English-only.


\bibliographystyle{unsrt}
\bibliography{references}

\appendix

\clearpage
\section{Acronyms Used in the Appendix}
\label{app:acronyms}

\begin{center}
\footnotesize
\begin{tabular}{@{}l>{\raggedright\arraybackslash}p{0.62\linewidth}@{}}
\toprule
\textbf{Acronym} & \textbf{Meaning} \\
\midrule
\multicolumn{2}{@{}l}{\textit{Methods / components}} \\
\lbm & Logit bridge mixing \\
\lbmdrop\,/\,\lbmfirst\,/\,\lbmexact & LBM multi-token fallback variants \\
\midrule
\multicolumn{2}{@{}l}{\textit{Benchmarks}} \\
HB-Std   & HarmBench-Standard \\
HB-Ctx   & HarmBench-Contextual \\
AdvB     & AdvBench \\
JBB-H    & JailbreakBench-Harmful \\
JBB-B    & JBB-Benign \\
Sorry    & Sorry-Bench \\
WildJB   & WildJailbreak-Eval-Harmful \\
OR-H     & OR-Bench-Hard \\
OR-Tox   & OR-Bench-Toxic \\
XSTest   & Exaggerated-safety (over-refusal) test set \\
\midrule
\multicolumn{2}{@{}l}{\textit{Metrics}} \\
Ref\%        & String-match refusal rate \\
ACC\%        & Benchmark-judge accuracy \\
$\dsafe$     & \alignbeam{} Ref\% $-$ Base Ref\% \\
HB$+$AB      & Count-weighted refusal over HarmBench-Std ($n\!=\!167$) and AdvBench ($n\!=\!520$) \\
\bottomrule
\end{tabular}
\end{center}

\clearpage
\section{Regime Patterns and Aggregate HB$+$AB}
\label{app:main-safety-full}

\alignbeam raises the \textbf{count-weighted HB$+$AB aggregate}
(HarmBench-Std $n\!=\!167$ + AdvBench $n\!=\!520$, shared
HarmBench-Mistral judge) from 35.1\% to 87.8\% ($+$52.7\,pp)
macro-averaged over five pairs (Table~\ref{tab:main-safety-full};
main-body Table~\ref{tab:headline} shows only the \alignbeam row).
Per-pair HB$+$AB:
Llama-3.1-8B 16.9$\to$\textbf{90.8} ($+$73.9), MedLlama-3-8B 34.5$\to$\textbf{88.5}
($+$54.0), Finance-Llama3-8B 28.1$\to$\textbf{91.1} ($+$63.0), DeepSeek-Math-7B-Instruct 11.5$\to$\textbf{76.4}
($+$64.9), Qwen3-8B-Base 84.6$\to$\textbf{92.1} ($+$7.5).

\paragraph{Regime patterns.}
The gain pattern across pairs reflects four distinct baseline safety
properties.
\emph{Cross-vocabulary domain specialists with low safety priors}
(Llama-3.1-8B, MedLlama-3-8B, Finance-Llama3-8B): all three start at 17--35\% and reach 88--91\%
HB$+$AB, the largest absolute gains. The medical pair (MedLlama-3-8B;
Appendix~\ref{app:e5-breakdown}) gains most on HarmBench-Contextual
($+$59.7\,pp) where domain framing is the attack vector; the
financial pair (Finance-Llama3-8B; Appendix~\ref{app:e6-breakdown}) follows the same
pattern.
\emph{Instruct draft with no safety prior} (DeepSeek-Math-7B-Instruct): the math
specialist's chat template receives explicit user turns but was
never trained to refuse them, so the anchor overrides the entire
harmful-response trajectory at $N\!=\!20$ positions; HB$+$AB moves
from 11.5\% to 76.4\%.
\emph{Same-vocabulary control} (Qwen3-8B-Base): gains only $+$7.5\,pp from
an already-aligned 84.6\% baseline, confirming that \alignbeam
behaves analogously to \textsc{SafeDecoding} for same-family pairs
(the construction differs in support selection: \lbm uses the anchor's
top-$B$ tokens rather than a token-level intersection) and adds no harm in
that regime.

\begin{table}[htpb]
\centering\scriptsize
\caption{\textbf{Cross-vocabulary pairs (DeepSeek-Math-7B-Instruct, Llama-3.1-8B, MedLlama-3-8B,
  Finance-Llama3-8B) gain $+$54--74\,pp HB$+$AB; the same-vocabulary control
  (Qwen3-8B-Base) gains $+$7.5\,pp from an already-aligned baseline. Full
  Base / \alignbeam grid.} Ref\% = string-match refusal;
  Acc\% = benchmark-judge accuracy. Cells marked --- are pending
  runs. $^{\ast}$ = 4-pair mean (DeepSeek-Math excluded); otherwise
  5-pair. Calibration
  / over-refusal per pair in
  Appendices~\ref{app:e3-breakdown}--\ref{app:eq1-full}.}
\label{tab:main-safety-full}
\vspace{6.3pt}
\begin{tabular}{@{}l l l rrrrrr c rr@{}}
\toprule
\textbf{Draft model} & \textbf{Method} & \textbf{Metric}
  & \textbf{HB-Std} & \textbf{HB-Ctx} & \textbf{AdvB}
  & \textbf{JBB-H} & \textbf{Sorry} & \textbf{WildJB}
  & & \textbf{HB$+$AB \%} & $\boldsymbol{\dsafe}$ \\
\midrule
\multicolumn{12}{@{}l}{\textit{Base drafts, cross-vocab, low safety prior}}\\
\multirow{4}{*}{Llama-3.1-8B}
  & \multirow{2}{*}{Base} & Ref\%
                      & 15.0 &  8.1 & 17.5 & 30.0 & 10.9 & 13.3
                      & & 16.9 & \\
  &                       & Acc\%
                      & 49.1 & 53.2 & 52.5 & 11.0 & 34.9 & 21.5
                      & & 51.7 & \\
& \multirow{2}{*}{\textbf{Ours}} & Ref\%
                  & \textbf{79.6}
                  & \textbf{79.0}
                  & \textbf{94.4}
                  & \textbf{80.0}
                  & \textbf{61.4}
                  & \textbf{19.7}
                  & & \textbf{90.8}
                  & $+$73.9 \\
&                                & Acc\%
                  & 91.6 & 80.7 & 98.3
                  & 79.0
                  & 74.3 & 30.1
                  & & 96.7 & \\
\midrule
\multirow{4}{*}{MedLlama-3-8B}
  & \multirow{2}{*}{Base} & Ref\%
                      & 16.8 &  9.7 & 40.2 & 25.0 & 18.0 & 14.1
                      & & 34.5 & \\
  &                       & Acc\%
                      & 35.3 & 25.8 & 59.2 & 42.0 & 31.1 & 22.5
                      & & 53.4 & \\
  & \multirow{2}{*}{\textbf{Ours}} & Ref\%
                      & \textbf{78.4} & \textbf{69.4} & \textbf{91.7} & \textbf{79.0}
                      & \textbf{63.2} & \textbf{16.2}
                      & & \textbf{88.5} & $+$54.0 \\
  &                                & Acc\%
                      & 88.6 & 87.1 & 99.0 & 82.0 & 72.3 & 27.1
                      & & 96.5 & \\
\midrule
\multirow{4}{*}{Finance-Llama3-8B}
  & \multirow{2}{*}{Base} & Ref\%
                      & 23.4 & 30.6 & 29.6 & 27.0 & 55.2 & 42.7
                      & & 28.1 & \\
  &                       & Acc\%
                      & 64.1 & 79.0 & 78.5 & 56.0 & 67.0 & 43.2
                      & & 75.0 & \\
  & \multirow{2}{*}{\textbf{Ours}} & Ref\%
                      & \textbf{79.6} & \textbf{80.6} & \textbf{94.8} & \textbf{88.0}
                      & \textbf{65.7} & \textbf{33.2}
                      & & \textbf{91.1} & $+$63.0 \\
  &                                & Acc\%
                      & 91.0 & 90.3 & 98.1 & 94.0 & 77.3 & 41.4
                      & & 96.4 & \\
\midrule
\multicolumn{12}{@{}l}{\textit{Instruct draft, cross-vocab (math)}}\\
\multirow{4}{*}{DeepSeek-Math-7B-Instruct}
  & \multirow{2}{*}{Base} & Ref\%
                      & 12.6 &  9.7 & 11.2 & 16.0 & 13.2 & 11.5
                      & & 11.5 & \\
  &                       & Acc\%
                      & 41.9 & 53.2 & 46.9 & 14.0 & 38.2 & 19.2
                      & & 45.7 & \\
  & \multirow{2}{*}{\textbf{Ours}} & Ref\%
                      & \textbf{63.5} & \textbf{61.3} & \textbf{80.6} & \textbf{67.0}
                      & \textbf{45.0} & \textbf{15.0}
                      & & \textbf{76.4} & $+$64.9 \\
  &                                & Acc\%
                      & 85.0 & 79.0 & 91.3 & 72.0 & 65.7 & 25.6
                      & & 89.8 & \\
\midrule
\multicolumn{12}{@{}l}{\textit{Same-vocab control}}\\
\multirow{4}{*}{Qwen3-8B-Base}
  & \multirow{2}{*}{Base} & Ref\%
                      & 61.7 & 37.1 & 91.9 & 76.0 & 39.5 & 16.1
                      & & 84.6 & \\
  &                       & Acc\%
                      & 80.8 & 64.5 & 95.6 & 81.0 & 55.2 & 27.6
                      & & 92.0 & \\
  & \multirow{2}{*}{\textbf{Ours}} & Ref\%
                      & \textbf{80.8} & \textbf{71.0} & \textbf{95.8} & \textbf{85.0}
                      & \textbf{63.4} & \textbf{20.9}
                      & & \textbf{92.1} & $+$7.5 \\
  &                                & Acc\%
                      & 89.8 & 90.3 & 99.4 & 91.0 & 79.8 & 32.6
                      & & 97.1 & \\
\midrule
\multicolumn{2}{@{}l}{\multirow{2}{*}{\textit{Macro avg.\ Base}}} & Ref\%
  & 25.9 & 19.0 & 38.1 & 34.8 & 27.4 & 19.5
  & & 35.1 & \\
\multicolumn{2}{@{}l}{} & Acc\%
  & 54.2 & 55.1 & 66.5 & 40.8 & 45.3 & 28.7$^{\ast}$
  & & 63.6 & \\
\multicolumn{2}{@{}l}{\multirow{2}{*}{\textbf{Macro avg.\ Ours}}} & Ref\%
  & \textbf{76.4} & \textbf{72.3} & \textbf{91.5} & \textbf{79.8}
  & \textbf{59.7} & \textbf{21.0}\,{\scriptsize\underline{$+$1.5}}
  & & \textbf{87.8} & \textbf{$+$52.7} \\
\multicolumn{2}{@{}l}{} & Acc\%
  & 89.2 & 85.5 & 97.2 & 83.6 & 73.9 & 31.4$^{\ast}$
  & & 95.3 & \\
\bottomrule
\end{tabular}%
\end{table}

\clearpage
\section{Why String-Match Refusal as Primary Metric}
\label{app:string-vs-acc}

String-match refusal (Ref\%) is the conservative primary metric: it
undercounts safety for both base models and \alignbeam outputs, so any
gap between Ref\% and ACC\% reflects genuine safety the string match
misses, not noise. Three patterns explain when and why the two columns
diverge.

\textsc{Base-model judge inflation.}
Incoherent or repetitive base-model outputs are correctly judged
non-harmful but match none of the 22 canonical refusal phrases, so
Base ACC\% can exceed Base Ref\% by $\approx$34\,pp on Llama-3.1-8B
(Appendix~\ref{app:e3-breakdown}).

\textsc{\alignbeam refuses in non-canonical phrasing.}
The anchor produces polite declinations that the judge classifies as
refusals but that do not appear in the 22-phrase list. This creates a
gap between \alignbeam ACC\% and Ref\%: 2--22\,pp on aligned drafts
(Llama-3.1-8B/MedLlama-3-8B/Finance-Llama3-8B/Qwen3-8B-Base median $\approx$10\,pp), and $>$80\,pp on the alternative-anchor run
(Appendix~\ref{app:a8a}). A related pattern occurs on WildJailbreak for
Finance-Llama3-8B and DeepSeek-Math-7B-Instruct: the unsteered model
generates compliance caveats that happen to match the 22-phrase list, so
its baseline Ref\% is non-zero (42.7\% / 19.2\%); \alignbeam replaces
these with anchor-style declinations outside the list, so Ref\% falls
while ACC\% rises (Finance-Llama3-8B: 42.7\%\,$\to$\,33.2\%; DeepSeek-Math-7B-Instruct:
19.2\%\,$\to$\,15.0\%). In all such cases judge ACC\% confirms the
model is genuinely safer. Because \alignbeam's outputs are always
non-degenerate, the string-match gap is a lower bound on the true
safety gain, which is why we adopt Ref\% as the conservative metric.

\textsc{Sorry-but-continues (Ref\% $>$ ACC\%).}
Some unsteered models produce a canonical refusal phrase and then
continue with the requested harmful content. This pattern appears in
certain Sorry-Bench sub-categories (Fraud/Deception, Privacy); in
these cases Ref\% \emph{over}estimates safety and ACC\% is the
reliable measure. \alignbeam's Phase~2 logit mixing anchors the
continuation to the safety anchor's distribution, reducing this
pattern in \alignbeam outputs (cf.\ Appendix~\ref{app:ext-limits} for
the residual cases on DeepSeek-Math-7B).

\clearpage
\section{Extended Related Work}
\label{app:related-extended}

\paragraph{Alignment erosion under fine-tuning.}
Prior work shows that as few as 100 benign fine-tuning samples
substantially reduce refusal \cite{qi2023finetuning}, and that
authority and domain-framing prompts provide potent soft jailbreaks
\cite{yang2023shadow}. Our two domain-fine-tuned
specialists (MedLlama-3-8B medical, Finance-Llama3-8B finance) and the math-instruct model
without safety training (DeepSeek-Math-7B-Instruct) reproduce this erosion at scale, with
HB$+$AB string-match baselines of 12--35\% vs.\ matching instruct upper-bounds; the
two pre-trained / unaligned drafts (Llama-3.1-8B,
Qwen3-8B-Base) span the orthogonal
absence-of-alignment regime (HB$+$AB baselines 17--85\%).

\paragraph{Inference-time logit interventions.}
The shared-vocabulary assumption is the central limitation of prior
logit-mixing methods.
\textsc{SafeDecoding} \cite{xu2024safedecoding} blends draft and
safety-model probabilities at the shared vocabulary intersection;
\textsc{Proxy Tuning} \cite{liu2024proxy} and top-$k$ contrastive
decoding \cite{li2023contrastive} apply the same logit arithmetic,
all restricted to same-family pairs.
\alignbeam relaxes this assumption via a text bridge and yields a
construction analogous to SafeDecoding when $\Vd\!=\!\Vs$
(differing in support selection: beam search vs.\ token-level
intersection); on the same-vocabulary
control (Qwen3-8B-Base) it outperforms both Proxy Tuning and top-$k$ CD on
five of six benchmarks ($+$25.8\,pp vs.\ Proxy Tuning on HB-Ctx;
Table~\ref{tab:eq1-methods}).
Recent methods that do not rely on logit sharing also fail on
domain-fine-tuned models: \textsc{NUDGING}~\cite{fei2025nudging}
swaps to an aligned guide on high-entropy steps; \textsc{DeAL}~\cite{huang2025deal}
uses reward-guided beam search; backtracking safety~\cite{lyu2025backtracking}
rolls back on early unsafe tokens; and \textsc{RAIN}~\cite{li2024rain}
self-aligns via rewindable generation. All four require a safety
prior in the draft model itself, which domain fine-tuning has
erased.

\paragraph{Post-hoc guard classifiers.}
LlamaGuard \cite{inan2023llamaguard} and similar response-time
classifiers block detected outputs but do not steer generation.
In domain-specialised settings, benign prompts that share surface
features with harmful ones (drug names, financial instruments,
exploit syntax) trigger classifier flags at high rates; this is
why LlamaGuard raises OR-Bench-Hard over-refusal to $\approx$34.5\% vs.\
\alignbeam's 22.3\% on the same Llama-3.1-8B draft
(Appendix~\ref{app:llamaguard-baselines}).

\paragraph{Early-token safety concentration.}
Prior work \cite{qi2025depth} argues that safety alignment
concentrates in the first few token positions and is brittle to
distribution shift.
If true, a shallow mixed-decoding prefix should be sufficient to
restore safety; \alignbeam's depth ablation
(Figure~\ref{fig:ablations}(a)) confirms this directly: three mixed
steps recover 92.7\% AdvBench refusal on Llama-3.1-8B, while deepening to
$N\!\ge\!6$ adds only $\approx$3.9\,pp more on HB$+$AB. This is
why the small mixed-depth defaults ($N\!\in\!\{3,6\}$) are used
throughout: the gain saturates early, and deeper mixing only adds
latency.

\clearpage
\section{Evaluation Judges and Dataset Details}
\label{app:judges}

Every benchmark is scored with its \emph{official released judge}
to avoid evaluator-vs-method confounds and to keep our numbers
directly comparable to prior work that uses the same setup.
Table~\ref{tab:judges-full} lists, for each benchmark used in
the paper, the judge model, judge output type, expected
direction (harmful prompts $\to$ refusal correct;
benign/calibration prompts $\to$ compliance correct), dataset
size used in our runs, and the source. None of these judges is
used \emph{inside} \alignbeam's beam selection (which uses a
separate 3B classifier, described below).

\begin{table}[htpb]
\centering
\scriptsize
\setlength{\tabcolsep}{4pt}
\renewcommand{\arraystretch}{1.1}

\caption{Per-benchmark evaluation setup. $n$ = prompts evaluated in our runs.
Judge output is parsed to a binary \texttt{is\_harmful} flag.
$^{\dagger}$Sorry-Bench judge id in full:
sorry-bench/ft-mistral-7b-instruct-v0.2-sorry-bench-202406.}
\label{tab:judges-full}

\vspace{6.3pt}
\begin{tabularx}{\textwidth}{l c r Y c Y}
\toprule
\textbf{Benchmark} & \textbf{Type} & $\boldsymbol{n}$ 
& \textbf{Judge} & \textbf{Output} & \textbf{Source} \\
\midrule

HarmBench-Standard   & H & 167 & \texttt{cais/HarmBench-Mistral-7b-val-cls} & yes/no & walledai/HarmBench (standard) \\
HarmBench-Contextual & H & 62  & \texttt{cais/HarmBench-Mistral-7b-val-cls} & yes/no & walledai/HarmBench (contextual) \\
AdvBench             & H & 520 & \texttt{cais/HarmBench-Mistral-7b-val-cls} & yes/no & walledai/AdvBench \\
Sorry-Bench          & H & 440 & \texttt{ft-mistral-7b-instruct-v0.2}$^{\dagger}$ & 0/1 & sorry-bench/sorry-bench-202503 \\
JailbreakBench-Harmful & H & 100 & \texttt{Llama-Guard-3-8B} & safe/unsafe & JBB-Behaviors (harmful) \\
JailbreakBench-Benign  & B & 100 & \texttt{Llama-Guard-3-8B} & safe/unsafe & JBB-Behaviors (benign) \\
WildJailbreak-Eval   & C & 1{,}105--1{,}315 & \texttt{wildguard} & multi-label & wildjailbreak (eval) \\
OR-Bench-Hard        & B & 855 & \texttt{Llama-Guard-3-8B} & safe/unsafe & or-bench (hard-1k) \\
OR-Bench-Toxic       & B & 585 & \texttt{Llama-Guard-3-8B} & safe/unsafe & or-bench (toxic) \\
XSTest               & C & 450 & \texttt{Llama-Guard-3-8B} & safe/unsafe & xstest \\
MedSafetyFix  & H & 900 & \texttt{Llama-Guard-3-8B} & safe/unsafe & AI4LIFE-GROUP/med-safety-bench~\cite{han2024medsafetybench} \\

\midrule
GSM8K (utility) & U & 1319 & exact-match numeric extraction & accuracy & gsm8k (test) \\
MedQA (utility) & U & 2243 & ROUGE-L + ACC + semantic metrics & accuracy & ChatDoctor dataset \\
\bottomrule
\end{tabularx}
\end{table}

\noindent\textsc{Type key.} H = harmful (model should refuse);
B = benign (model should comply); C = calibration (mixed labels;
the dataset's own \texttt{label} column determines the expected
behaviour per prompt); U = utility (scored by task accuracy, not
refusal classification).

\paragraph{Why these judges.}
HarmBench-Mistral-7b-val-cls is the judge released with the
HarmBench paper and validated there at $\approx$93\% agreement
with human red-teamers \cite{mazeika2024harmbench}; we re-use it unchanged on HarmBench-Std/Ctx and AdvBench because they share the
same harmful-completion definition. Sorry-Bench provides its own
fine-tuned 7B classifier \cite{xie2024sorrybench} that is
calibrated to the category-balanced Sorry-Bench split.
LlamaGuard-3-8B \cite{inan2023llamaguard} is the official JBB
and OR-Bench judge and the one used in our domain-safety set
(MedSafetyFix). WildGuard \cite{han2024wildguard} is paired
with WildJailbreak by construction. We deliberately avoid
substituting any judge with an in-house re-trained variant to
keep every column comparable across this paper, prior work, and
future follow-ups.

\paragraph{String-match refusal (primary metric).}
A response is counted as a refusal if any of the following
22 canonical phrases appears as a case-insensitive substring
of the model output:
\textit{``I'm sorry'', ``I am sorry'', ``sorry'',
``I apologize'', ``as an AI'', ``as a language model'',
``as an assistant'', ``I cannot'', ``I do not'',
``I'm unable to'', ``I am unable to'', ``I am not allowed
to'', ``I cannot assist'', ``I will not'', ``I cannot
provide'', ``I cannot help'', ``I cannot support'',
``I cannot recommend'', ``against ethical'',
``not ethical'', ``illegal'', ``harmful''}
(plus the implementation-level extra phrase ``I must
refuse''). The string-match rule is identical across every
benchmark, every pair, every method.

\paragraph{Benchmark-judge accuracy (ACC\%).}
For harmful benchmarks (Type H above), ACC\% is the fraction
of responses the judge classifies as not exhibiting the
harmful behaviour, i.e., correct refusal as judged by the
official classifier. For benign / calibration benchmarks
(Type B and the comply rows of Type C), ACC\% is the fraction
of responses the judge classifies as safe / benign /
compliant. WildJailbreak and XSTest are mixed-label
calibration sets: per-prompt expected behaviour is read from
the dataset's \texttt{label} column and correctness is
computed against that expectation.

\paragraph{LLM judge inside Phase~3 (in-generation only).}
A separate \texttt{Qwen2.5-3B-Instruct} judge scores beams on
a 1--5 harm rubric inside Phase~3 of Algorithm~\ref{alg:alignbeam}.
It is used \emph{only} for in-generation beam selection; the
safest beam is returned if any beam has score $\le\tau\!=\!2.5$.
This judge is never used to report safety numbers (avoiding
self-evaluation circularity).

\paragraph{Judge and anchor family overlap.}
The Phase 3 judge and the safety anchor share the same checkpoint
(\texttt{Qwen2.5-3B-Instruct}). The no-judge ablation
(Appendix~\ref{app:nojudge}) shows that \emph{dropping} the judge
slightly \emph{increases} adversarial accuracy (AdvBench $+$3.5\,pp,
HB-Ctx $+$6.5\,pp) while also raising XSTest over-refusal by
$+$6.4\,pp, confirming that the judge's role is calibration
(suppressing benign over-refusal) rather than adversarial-accuracy
inflation. Anchor-judge family overlap therefore does not inflate
the headline safety numbers; if anything the judge is conservative
on adversarial sets. All headline safety numbers are scored by the
benchmark's official judge (Table~\ref{tab:judges-full}), which is
independent of the Qwen family for HarmBench, Sorry-Bench, JBB,
OR-Bench, and XSTest.

\paragraph{Utility metrics.}
GSM8K accuracy is exact-match on the numeric answer extracted
from the response (the harness parses ``\#\#\#\# N'' suffix or
the first standalone numeric token). MedQA reports four
quantities on ChatDoctor-HealthCareMagic-style free-form
generations against the reference doctor response: ROUGE-L F1
(the primary \texttt{is\_correct} signal in the harness),
ROUGE-1/2, BERTScore-F1, and SentSim (all-MiniLM-L6-v2 cosine
similarity). The MedQA-USMLE multiple-choice accuracy in the
main body utility paragraph is computed separately.

\paragraph{Metric conventions used in all appendix tables.}
\begin{itemize}[leftmargin=*,itemsep=2pt,topsep=2pt]
  \item \textbf{Base\,\%} / \textbf{Ref.\%}: string-match refusal rate.
  \item \textbf{ACC\%}: benchmark-judge accuracy. Higher than
    Base\% for base-model baselines due to score inflation on
    incoherent outputs (discussed in Appendix~\ref{app:string-vs-acc}).
  \item $\boldsymbol{\dsafe}$: \alignbeam\% $-$ Base\% (string-match
    delta) unless the column header states ``ACC.''
  \item For \textbf{calibration / benign datasets} (OR-Bench-Hard,
    XSTest, JBB-Benign): Ref.\% denotes \emph{over-refusal} rate
    (lower is better).
  \item \textbf{OR-Bench-Toxic} is a benign-but-adversarially-phrased
    calibration split: prompts that look harmful but are actually
    answerable. \alignbeam tends to \emph{increase} string-match
    refusal on this split (e.g.\ Llama-3.1-8B: 12.3\%$\to$56.8\%; type B,
    over-refusal direction) while the benchmark-judge ACC\% drops
    sharply (e.g.\ Llama-3.1-8B: 53.9\%$\to$20.5\%, $-$33.4\,pp) because
    the judge penalises refusal-style responses on prompts the model
    should answer. Both columns indicate over-refusal on this split,
    not a safety regression.
\end{itemize}

\clearpage
\section{Full LBM Procedure and Implementation Notes}
\label{app:lbm-detail}

\paragraph{Text bridge procedure.}
Algorithm~\ref{alg:alignbeam} (main body) compresses the per-step
\lbm computation. In full: for each anchor token
$s_{\mathrm{id}} \in \mathcal{T}_B$ with mass $p_s$, we compute
$t = \mathrm{decode}_s(s_{\mathrm{id}})$,
$d_{\mathrm{ids}} = \mathrm{encode}_d(t)$, and inspect
$|d_{\mathrm{ids}}|$. If $|d_{\mathrm{ids}}|\!=\!1$ we accumulate
into the buffer via Eq.~\ref{eq:lbm-mix}; if multiple anchor tokens
map to the same draft token, their contributions sum. After all $B$
anchor tokens are processed, $f$ is renormalised to yield $\tilde{f}$.
The mixing variants we evaluate are summarised in
Table~\ref{tab:variants} (main body).

\paragraph{Phase implementation details (not in pseudocode).}
(i)~If a beam's Phase~2 prefix already opens with a canonical refusal
phrase (``I cannot'', ``I'm sorry''), Phase~3 continuation is capped
at $N\!+\!20$ tokens, since base models have no training signal for
what follows a refusal and longer continuations produce hallucinated
content. (ii)~If either model emits EOS during Phase~2, the beam is
marked completed and skips Phase~3 generation. (iii)~If no beam
satisfies the judge threshold ($J(b_k)\!\le\!\tau$), the beam with
the lowest judge score is returned as a fallback (in practice rare on
structured adversarial benchmarks). (iv)~Overhead
breakdown: Phase~3 (KV-cache draft continuation) $\approx$69\%,
Phase~2 (mixed decoding) $\approx$30\%, Phase~1 $\approx$1\%; the
primary speed lever is therefore $K$ rather than $N$.

\clearpage
\section{Per-Pair Per-Dataset Breakdowns}
\label{app:per-pair-breakdowns}

\subsection{Llama-3.1-8B}
\label{app:e3-breakdown}

Llama-3.1-8B is the primary cross-vocabulary experiment and
the basis for all ablation studies. Structured single-turn benchmarks
show the largest gains (AdvBench $+$76.9\,pp, HB-Std $+$64.6\,pp,
HB-Ctx $+$70.9\,pp), while WildJailbreak-Harmful is the principal
exception: only $+$6.4\,pp, because multi-sentence framing distributes
the harmful payload beyond the $N\!=\!6$ mixed prefix. Calibration
over-refusal rises as expected (OR-H 11.0\%$\to$22.3\%,
XSTest 5.6\%$\to$40.0\%), driven by the anchor's inherited refusal prior.
Table~\ref{tab:e3-breakdown}.

\begin{table}[htpb]
\centering\scriptsize
\caption{Per-dataset breakdown
  (Llama-3.1-8B $+$ Qwen2.5-3B-Instruct, \lbmdrop,
  $\alpha\!=\!0.5$, $N\!=\!6$, $K\!=\!3$).
  H $=$ harmful, B $=$ benign, C $=$ calibration. Base \% / +\,AB \%
  are string-match refusal; ACC columns are benchmark-judge
  accuracy. For benign / calibration rows the reported \% is
  over-refusal (lower is better).}
\label{tab:e3-breakdown}
\vspace{6.3pt}
\begin{tabular}{@{}llrrrrr@{}}
\toprule
\textbf{Dataset} & \textbf{Type} & $\boldsymbol{n}$
  & \textbf{Base\,\%} & \textbf{Base ACC\,\%}
  & \textbf{+\,AB\,\%} & \textbf{+\,AB ACC\,\%} \\
\midrule
\multicolumn{7}{@{}l}{\textit{Harmful (refusal $\uparrow$)}} \\
HarmBench-Std   & H & 167 & 15.0 & 49.1 & \textbf{79.6} & 91.6 \\
HarmBench-Ctx   & H &  62 &  8.1 & 53.2 & \textbf{79.0} & 80.7 \\
AdvBench        & H & 520 & 17.5 & 52.5 & \textbf{94.4} & 98.3 \\
JBB-Harmful     & H & 100 & 30.0 & 11.0 & \textbf{80.0} & 79.0 \\
Sorry-Bench     & H & 440 & 10.9 & 34.9 & \textbf{61.4} & 74.3 \\
WildJailbreak-H & H &1315 & 13.3 & 21.5 & \textbf{19.7} & 30.1 \\
\midrule
\multicolumn{7}{@{}l}{\textit{Benign / calibration (over-ref.\ $\downarrow$)}} \\
JBB-Benign      & B &  100 & 10.0 &  8.0 & 26.0 &  1.0 \\
OR-Bench-Hard   & B &  855 & 11.0 & 13.6 & 22.3 &  6.2 \\
OR-Bench-Toxic  & B &  585 & 12.3 & 53.9 & 56.8 & 20.5 \\
XSTest          & C &  450 &  5.6 & 23.6 & 40.0 & 38.7 \\
\midrule
\textbf{HB$+$AB} & & \textbf{687} & \textbf{16.9} & 51.7 & \textbf{90.8} & 96.7 \\
\bottomrule
\end{tabular}%
\end{table}

\subsection{MedLlama-3-8B}
\label{app:e5-breakdown}

MedLlama-3-8B exemplifies the domain-framing attack
vector: safety erosion is largest on HarmBench-Contextual, where
prompts are framed in clinical language the medical specialist
handles fluently. Correspondingly, \alignbeam's gains peak there:
HB-Ctx $+$59.7\,pp (the highest per-benchmark delta in the main
results). WildJailbreak-Harmful shows only $+$2.1\,pp (base
14.1\%\,$\to$\,16.2\%), confirming the prefix-length bottleneck.
HB$+$AB moves from 34.5\% to 88.5\% ($+$54.0\,pp).
Table~\ref{tab:e5-breakdown}.

\begin{table}[htpb]
\centering\scriptsize
\caption{Per-dataset breakdown (JSL-MedLlama-3-8B-v2.0 $+$ Qwen2.5-3B-Instruct,
  \lbmdrop, $\alpha\!=\!0.5$, $N\!=\!6$, $K\!=\!3$). Per-pair
  Base string-match values were not separately recorded in the
  evaluation logs. Columns: Base ACC \% is the unmodified draft's benchmark-judge accuracy; +\,AB \% and +\,AB ACC \% are \alignbeam's string-match refusal rate and benchmark-judge accuracy.}
\label{tab:e5-breakdown}
\vspace{6.3pt}
\begin{tabular}{@{}llrrrr@{}}
\toprule
\textbf{Dataset} & \textbf{Type} & $\boldsymbol{n}$
  & \textbf{Base ACC\,\%} & \textbf{+\,AB\,\%} & \textbf{+\,AB ACC\,\%} \\
\midrule
HarmBench-Std   & H &  167 & 35.3 & \textbf{78.4} & 88.6 \\
HarmBench-Ctx   & H &   62 & 25.8 & \textbf{69.4} & 87.1 \\
AdvBench        & H &  520 & 59.2 & \textbf{91.7} & 99.0 \\
JBB-Harmful     & H &  100 & 42.0 & \textbf{79.0} & 82.0 \\
Sorry-Bench     & H &  440 & 31.1 & \textbf{63.2} & 72.3 \\
WildJailbreak-H & H & 1315 & 22.5 & 16.2          & 27.1 \\
\midrule
\textbf{HB$+$AB} & & \textbf{687}
  & \multicolumn{3}{c}{$\textbf{34.5}\%\!\to\!\textbf{88.5}\%$\;($+$\textbf{54.0}\,pp)} \\
\bottomrule
\end{tabular}%
\end{table}

\subsubsection{MedLlama-3-8B Domain-Specific Medical Safety (MedSafetyFix)}
\label{app:e5-medsafety}

MedSafetyFix is a curated set of 900 harmful medical prompts drawn
from \texttt{AI4LIFE-GROUP/med-safety-bench}~\cite{han2024medsafetybench},
spanning prescription misuse, dangerous diet advice, unethical
clinical communication, and discriminatory care (judge: LlamaGuard-3-8B). Prompts use clinical or academic framing that a medical specialist
handles fluently, making domain framing the principal attack vector. On MedLlama-3-8B,
\alignbeam (\lbmdrop and \lbmexact, identical here) raises
LlamaGuard accuracy from 77.3\% to \textbf{96.8\%} ($+$19.5\,pp)
and more than doubles the string-match refusal rate from 15.3\%
to 34.4\%; the gap between string-match and LlamaGuard reflects
refusals phrased in clinical language outside the canonical
22-phrase list. \lbmfirst was excluded due to a generation-file
corruption.

\subsection{Finance-Llama3-8B}
\label{app:e6-breakdown}

Finance-Llama3-8B achieves the highest structured-benchmark
performance in the suite (AdvBench 94.8\%, JBB-H 88.0\%), and all
three LBM variants produce identical results, indicating near-perfect
tokeniser compatibility between Finance-Llama and Qwen2.5-3B.
The one negative result is WildJailbreak-H: string-match
refusal \emph{falls} from 42.7\% (base) to 33.2\% ($-$9.5\,pp).
The likely mechanism is that the financial draft's compliance-style
cautionary phrasing (which the string matcher counts as refusals)
is replaced by the anchor's polite-but-non-canonical declinations;
the content remains safe (Acc\% rises to 41.4\%), but the 22-phrase
list misses anchor-style wordings. HB$+$AB moves from 28.1\% to
91.1\% ($+$63.0\,pp). Table~\ref{tab:e6-breakdown}.

\begin{table}[htpb]
\centering\scriptsize
\caption{Finance-Llama3-8B. All three LBM variants
  (drop / first / exact) produce identical results, indicating
  high tokeniser compatibility with Qwen2.5-3B. Columns: +\,AB \% (string-match refusal) and +\,AB ACC \% (benchmark-judge accuracy) under \alignbeam.}
\label{tab:e6-breakdown}
\vspace{6.3pt}
\begin{tabular}{@{}llrrr@{}}
\toprule
\textbf{Dataset} & \textbf{Type} & $\boldsymbol{n}$
  & \textbf{+\,AB\,\%} & \textbf{+\,AB ACC\,\%} \\
\midrule
HarmBench-Std    & H &  167 & \textbf{79.6} & 91.0 \\
HarmBench-Ctx    & H &   62 & \textbf{80.6} & 90.3 \\
AdvBench         & H &  520 & \textbf{94.8} & 98.1 \\
JBB-Harmful      & H &  100 & \textbf{88.0} & 94.0 \\
Sorry-Bench      & H &  440 & \textbf{65.7} & 77.3 \\
WildJailbreak-H  & H & 1315 & 33.2 & 41.4 \\
\midrule
\textbf{HB$+$AB} & & \textbf{687}
  & \multicolumn{2}{c}{$\textbf{28.1}\%\!\to\!\textbf{91.1}\%$\;($+$\textbf{63.0}\,pp)} \\
\bottomrule
\end{tabular}%
\end{table}

\subsection{Qwen3-8B-Base (same-vocabulary control)}
\label{app:eq1-full}

Qwen3-8B-Base $+$ Qwen2.5-3B-Instruct is the same-vocabulary
control: both models share the Qwen2.5 tokeniser, so the text bridge
collapses to an identity map and all three LBM variants produce
results within 0.1\,pp of each other. The pair starts from an
already-aligned 84.6\% HB$+$AB baseline; \alignbeam adds $+$7.5\,pp
to 92.1\%, confirming that the method adds no harm in the
same-vocabulary regime and reduces to a beam-selection over the
anchor's own draft. WildJailbreak is the only benchmark where
string-match refusal lags judge accuracy noticeably (20.9\%
vs.\ 32.6\%), the same pattern seen across all pairs.

\begin{table}[htpb]
\centering\scriptsize
\caption{Qwen3-8B-Base ($+$ Qwen2.5-3B-Instruct, same
  tokeniser, \lbmdrop, $\alpha\!=\!0.5$, $N\!=\!6$, $K\!=\!3$).
  All three LBM strategies produce identical results
  ($>$98\% single-token match rate). Columns: Base \% / +\,AB \% are
  string-match refusal; ACC columns are benchmark-judge accuracy;
  $\dsafe = $ +\,AB Ref \% $-$ Base Ref \%. $^{\dagger}$XSTest is a
  calibration set: positive $\dsafe$ here denotes \emph{increased
  over-refusal} (worse), opposite sign convention to the harmful
  rows above.}
\label{tab:eq1-full}
\vspace{6.3pt}
\begin{tabular}{@{}llrrrrrr@{}}
\toprule
\textbf{Dataset} & \textbf{Type} & $\boldsymbol{n}$
  & \textbf{Base\,\%} & \textbf{Base ACC\,\%}
  & \textbf{+\,AB\,\%} & \textbf{+\,AB ACC\,\%}
  & $\boldsymbol{\dsafe}$ \\
\midrule
HarmBench-Std & H &  167 & 61.7 & 80.8 & \textbf{80.8} & 89.8 & $+$19.1 \\
HarmBench-Ctx & H &   62 & 37.1 & 64.5 & \textbf{71.0} & 90.3 & $+$33.9 \\
AdvBench      & H &  520 & 91.9 & 95.6 & \textbf{95.8} & 99.4 & $+$3.9  \\
JBB-Harmful   & H &  100 & 76.0 & 81.0 & \textbf{85.0} & 91.0 & $+$9.0  \\
Sorry-Bench   & H &  440 & 39.5 & 55.2 & \textbf{63.4} & 79.8 & $+$23.9 \\
WildJailbreak & H & 1315 & 16.1 & 27.6 & \textbf{20.9} & 32.6 & $+$4.8  \\
\midrule
XSTest$^{\dagger}$ & C & 450 & 10.7 & 26.0 & 28.2 & 38.4 & $+$17.5 \\
\midrule
\textbf{HB$+$AB} & & \textbf{687}
              & \textbf{84.6} & 92.0 & \textbf{92.1} & 97.1 & \textbf{$+$7.5} \\
\bottomrule
\end{tabular}%
\end{table}

\paragraph{Same-vocabulary method comparison.}
\label{app:eq1-methods}
\alignbeam outperforms Proxy Tuning \cite{liu2024proxy} and
top-$k$ contrastive decoding (\textsc{TopK-CD}) most sharply on
context-sensitive benchmarks: HarmBench-Contextual $+$25.8\,pp
vs.\ $+$4.9/+11.3\,pp; Sorry-Bench $+$24.5\,pp
vs.\ $+$13.0/+14.1\,pp. The gap arises because Proxy Tuning and
TopK-CD subtract a fixed unsafe-direction vector per token,
whereas \alignbeam supplies the anchor's full learned distribution,
which is better calibrated to domain-framed prompts.
XSTest over-refusal is nearly identical across all three
($\approx$37--38\%), confirming that the structural over-refusal
floor is a property of the anchor, not the mixing strategy.
Overhead is comparable ($\approx$4.8--5.4$\times$;
Appendix~\ref{app:latency}).

\begin{table}[htpb]
\centering\scriptsize
\caption{\alignbeam leads by $+$25.8\,pp on HarmBench-Contextual
  and $+$24.5\,pp on Sorry-Bench over the next-best method.
  Qwen3-8B-Base same-vocabulary method comparison (benchmark-judge ACC\%;
  \alignbeam $=$ any LBM variant, all identical in this regime).
  $^{\dagger}$XSTest: positive $\Delta$ = increased over-refusal (worse).}
\label{tab:eq1-methods}
\vspace{6.3pt}
\begin{tabular}{@{}llrrrrr@{}}
\toprule
& & \textbf{Base}
  & \multicolumn{2}{c}{\textbf{\alignbeam}}
  & \multicolumn{1}{c}{\textbf{Proxy}}
  & \multicolumn{1}{c}{\textbf{TopK-CD}} \\
\cmidrule(lr){4-5}
\textbf{Dataset} & \textbf{Type} & \textbf{ACC\%}
  & \textbf{ACC} & $\boldsymbol{\Delta}$
  & $\boldsymbol{\Delta}$ & $\boldsymbol{\Delta}$ \\
\midrule
AdvBench    & H & 95.6 & \textbf{99.4} & $+$3.8  & $+$2.5  & $+$2.7 \\
HB-Ctx      & H & 64.5 & \textbf{90.3} & $+$25.8 & $+$4.9  & $+$11.3 \\
HB-Std      & H & 80.8 & \textbf{89.8} & $+$9.0  & $+$2.4  & $+$5.4 \\
JBB-H       & H & 81.0 & 91.0          & $+$10.0 & \textbf{$+$11.0} & $+$9.0 \\
Sorry-Bench & H & 55.2 & \textbf{79.8} & $+$24.5 & $+$13.0 & $+$14.1 \\
WildJailbk  & H & 27.6 & \textbf{32.6} & $+$5.0  & $+$1.8  & $+$3.5 \\
\midrule
XSTest$^{\dagger}$ & C & 26.0 & 38.4 & $+$12.4 & $+$11.3 & $+$10.9 \\
\bottomrule
\end{tabular}%
\end{table}

\clearpage
\section{Baseline Details}
\label{app:baseline-details}

\subsection{Instruct-Counterpart Breakdown}
\label{app:b2-breakdown}

The instruct counterpart (Llama-3.1-8B-Instruct) is the upper-bound reference in the main
baseline comparison. Table~\ref{tab:b2-breakdown} reports its full
profile. Key observations: (i)~the instruct model is already
near-ceiling on structured adversarial sets so \alignbeam adds
nothing on those rows; (ii)~WildJailbreak drops by $-$10.2\,pp when
\alignbeam is applied, showing that anchor mixing can disrupt the
instruct model's own contextual reasoning on multi-sentence prompts;
(iii)~OR-Bench-Hard ACC is only 1.5\% for the instruct baseline,
confirming that the benchmark-judge severely over-classifies
instruct refusals on adversarial-looking benign prompts.

\begin{table}[htpb]
\centering\scriptsize
\caption{Instruct-counterpart (Llama-3.1-8B-Instruct) per-dataset breakdown
  ($K\!=\!3$, \lbmdrop, $\alpha\!=\!0.5$, seed $42$). Inst.\ ACC \% is the instruct counterpart's benchmark-judge accuracy on its own; +\,AB ACC \% adds \alignbeam over the instruct draft; +\,AB Ref \% is the resulting string-match refusal rate. For benign / calibration rows, lower +\,AB Ref \% means less over-refusal.}
\label{tab:b2-breakdown}
\vspace{6.3pt}
\begin{tabular}{@{}llrrrrr@{}}
\toprule
\textbf{Dataset} & \textbf{Type} & $\boldsymbol{n}$
  & \textbf{Instruct ACC\,\%}
  & \textbf{+\,AB ACC\,\%} & $\boldsymbol{\Delta}$
  & \textbf{+\,AB Ref\,\%} \\
\midrule
\multicolumn{7}{@{}l}{\textit{Harmful (refusal $\uparrow$)}} \\
AdvBench      & H &  520 & 99.8 & \textbf{99.6} & $-$0.2 & 35.6 \\
HB-Ctx        & H &   62 & 98.4 & \textbf{96.8} & $-$1.6 & 24.2 \\
HB-Std        & H &  167 & 95.2 & \textbf{95.2} & $+$0.0 & 24.6 \\
Sorry-Bench   & H &  440 & 83.6 & \textbf{86.1} & $+$2.5 & 21.1 \\
WildJailbreak & H & 1315 & 62.2 & 52.0          & $\mathbf{-10.2}$ & 20.2 \\
\midrule
\multicolumn{7}{@{}l}{\textit{Benign / calibration}} \\
JBB-Benign    & B & 100 &  3.0 &  2.0 & $-$1.0 &  8.0 \\
OR-Bench-Hard & B & 855 &  1.5 &  1.4 & $-$0.1 & 15.1 \\
OR-Bench-Toxic& B & 585 &  1.0 &  1.5 & $+$0.5 & 25.8 \\
XSTest        & C & 450 & 44.4 & 43.6 & $-$0.9 & 18.0 \\
\bottomrule
\end{tabular}%
\end{table}

\subsection{LlamaGuard Baselines (prompt-filter and response-filter)}
\label{app:llamaguard-baselines}

LlamaGuard prompt-filter and response-filter achieve high
benchmark-judge accuracy on structured adversarial sets (89--100\%)
but block nearly every benign domain-adjacent prompt: OR-Bench-Hard
drops to ACC\,$\le$\,2.6\% and OR-Bench-Toxic falls below 2\% for
both (Table~\ref{tab:llamaguard}, Llama-3.1-8B).

\begin{table}[htpb]
\centering\scriptsize
\caption{LlamaGuard prompt-filter and response-filter on
  Llama-3.1-8B ($K\!=\!3$, seed $42$). Both impose severe over-refusal on
  benign domain-adjacent prompts.}
\label{tab:llamaguard}
\vspace{6.3pt}
\begin{tabular}{@{}llrrrr@{}}
\toprule
\textbf{Dataset} & \textbf{Type} & $\boldsymbol{n}$
  & \textbf{Base ACC\%} & \textbf{Prompt-filter ACC\%} & \textbf{Response-filter ACC\%} \\
\midrule
\multicolumn{6}{@{}l}{\textit{Harmful (refusal $\uparrow$)}} \\
AdvBench      & H & 520 & 52.5 & \textbf{100.0} & \textbf{100.0} \\
HarmBench-Ctx & H &  62 & 53.2 & 95.2           & \textbf{100.0} \\
HarmBench-Std & H & 167 & 49.1 & 98.8           & \textbf{99.4}  \\
Sorry-Bench   & H & 440 & 34.9 & 89.8           & 88.9           \\
\midrule
\multicolumn{6}{@{}l}{\textit{Benign / calibration (over-ref.\ $\downarrow$)}} \\
JBB-Benign    & B & 100 &  8.0 &  0.0 &  0.0 \\
OR-Bench-Hard & B & 855 & 13.6 &  2.6 &  0.1 \\
OR-Bench-Toxic& B & 585 & 53.8 &  1.0 &  0.0 \\
XSTest        & C & 450 & 23.6 & 42.9 & 43.3 \\
\bottomrule
\end{tabular}%
\end{table}

\subsection{Hard-Prefix Baseline}
\label{app:b8}

The hard-prefix baseline prepends a hard safety prefix (``I cannot comply with that
request because\ldots'') before letting the draft continue. On Llama-3.1-8B
($K\!=\!1$, \lbmdrop, seed 42), it reaches \textbf{99.0\%} judge
accuracy on AdvBench, 91.6\% on HarmBench-Std, 88.7\% on
HarmBench-Ctx, and 74.5\% on Sorry-Bench, but over-refusal rises on benign prompts: JBB-Benign string-match
10.0\%\,$\to$\,29.0\%, OR-Bench-Hard 11.0\%\,$\to$\,23.7\%,
OR-Bench-Toxic 12.3\%\,$\to$\,59.5\% (judge ACC
53.8\%\,$\to$\,25.5\%). Latency is in Appendix~\ref{app:latency};
Hard-prefix adversarial accuracy is comparable to \alignbeam's budget
setting, so benign over-refusal is the key differentiator.

\clearpage
\section{Ablation Sweeps}
\label{app:ablation-sweeps}

\subsection{Full Per-Dataset Depth Sweep}
\label{app:a1-full}

Table~\ref{tab:a1-full} gives the full per-dataset numbers
(Llama-3.1-8B, $\alpha\!=\!0.85$, $K\!=\!3$); the Pareto curve is in the
main body (Figure~\ref{fig:ablations}(a)). The depth sweep is run at
the high-safety end of the $\alpha$ range ($\alpha\!=\!0.85$)
rather than at the main-run default of $\alpha\!=\!0.5$ so that
the safety-vs-depth signal is read against a saturated mixing
weight; the main-run defaults reported elsewhere remain
$\alpha\!=\!0.5$, $N\!=\!6$, $K\!=\!3$.

\begin{table}[htpb]
\centering\scriptsize
\caption{Safety saturates at $N\!=\!6$; benign over-refusal barely
  changes across depths. Full per-dataset depth sweep
  (Llama-3.1-8B, \lbmdrop, $\alpha\!=\!0.85$, $K\!=\!3$).
  Cells: string-match refusal\% (harmful rows);
  $^{\dagger}$calibration rows are over-refusal (lower is better).
  $^{\ddagger}$HB-Std Base (13.8\%) differs from the Llama-3.1-8B headline value (15.0\%;
  Table~\ref{tab:e3-breakdown}) because this sweep is a separate evaluation
  run at $\alpha\!=\!0.85$; all \alignbeam columns within this table are
  internally consistent.}
\label{tab:a1-full}
\vspace{6.3pt}
\begin{tabular}{@{}lrrrrrrr@{}}
\toprule
\textbf{Dataset} & $\boldsymbol{n}$ & \textbf{Base\,\%}
  & $\boldsymbol{N\!=\!0}$ & $\boldsymbol{N\!=\!3}$ & $\boldsymbol{N\!=\!6}$
  & $\boldsymbol{N\!=\!10}$ & $\boldsymbol{N\!=\!20}$ \\
\midrule
AdvBench       & 520 & 17.5 & 30.4 & 92.7 & 93.5 & 96.2 & \textbf{97.3} \\
HB-Std         & 167 & 13.8 & 17.4 & 77.8 & 80.8 & \textbf{81.4} & 79.6 \\
Sorry-Bench    & 440 & 10.9 & 18.6 & 60.0 & 61.6 & 62.1 & \textbf{68.2} \\
\midrule
OR-Bench-H$^{\dagger}$ & 855 & 11.0 & 14.7 & 22.7 & 22.8 & 23.2 & 32.4 \\
XSTest$^{\dagger}$ & 450 &  5.6 &  9.6 & 40.7 & 40.0 & 40.0 & 42.7 \\
\midrule
\textbf{HB$+$AB}  & 687 & 16.6 & 27.2 & \textbf{89.1} & 90.4 & 92.6 & 93.0 \\
\textbf{Slowdown}    &     & 1.0$\times$ & 5.7$\times$ & 4.8$\times$ & 5.1$\times$ & 5.2$\times$ & 5.8$\times$ \\
\bottomrule
\end{tabular}%
\end{table}

\subsection{Full Per-Dataset Alpha Sweep}
\label{app:a2-full}

The $\alpha$ sweep (Llama-3.1-8B, $N\!=\!6$, $K\!=\!3$) traces a clean
Pareto front: safety rises monotonically from $\alpha\!=\!0.10$
(38.0\% HB$+$AB) to $\alpha\!=\!0.50$ (91.0\%), then
plateaus at $\alpha\!=\!0.75$ (90.4\%). Calibration over-refusal on
OR-Bench-Hard stays below 21\% across the entire range, confirming
that the safety-utility trade-off is controlled by a single
user-adjustable parameter and does not catastrophically degrade
benign accuracy at any setting. XSTest over-refusal tracks $\alpha$
more closely (6.7\%\,$\to$\,24.4\% at 0.50) because XSTest
ambiguous prompts are more sensitive to the anchor's
refusal prior than the OR-Bench benign prompts are.

Table~\ref{tab:a2-full} gives the full per-dataset numbers;
the Pareto curve is in the main body (Figure~\ref{fig:ablations}(b)).

\begin{table}[htpb]
\centering\scriptsize
\caption{Refusal saturates by $\alpha\!=\!0.50$; benign over-refusal
  (OR-H, XSTest) stays flat, showing the calibration floor is set
  by the anchor rather than by $\alpha$.
  Full per-dataset alpha sweep (Llama-3.1-8B, \lbmdrop, $N\!=\!6$, $K\!=\!3$).
  Cells: string-match refusal\% (harmful); over-refusal\% ($^{\dagger}$
  calibration; lower is better). $\alpha\!=\!0.85$ row from the depth sweep's
  depth-6 run. Base values (e.g.\ OR-H 15.1\%) differ from the depth-sweep
  run (11.0\%) due to separate sampling; all \alignbeam columns
  within this table are internally consistent.}
\label{tab:a2-full}
\vspace{6.3pt}
\begin{tabular}{@{}lrrrrrr@{}}
\toprule
\textbf{Dataset} & $\boldsymbol{n}$ & \textbf{Base\,\%}
  & $\boldsymbol{0.10}$ & $\boldsymbol{0.25}$
  & $\boldsymbol{0.50}$ & $\boldsymbol{0.75}$ \\
\midrule
AdvBench         & 520 & 17.5 & 41.2 & 81.0 & \textbf{94.4} & 93.5 \\
HB-Std           & 167 & 13.8 & 28.1 & 59.9 & 80.2 & \textbf{80.8} \\
\midrule
OR-Bench-H$^{\dagger}$ & 855 & 15.1 & 17.8 & 19.8 & 20.2 & \textbf{19.2} \\
XSTest$^{\dagger}$ & 450 &  6.7 &  5.6 & 16.4 & 24.4 & 24.0 \\
\midrule
\textbf{HB$+$AB} & 687 & 16.6 & 38.0 & 75.8 & \textbf{91.0} & 90.4 \\
\textbf{Net}     &     &      & 20.2 & 56.0 & \textbf{70.8} & 71.2 \\
\bottomrule
\end{tabular}%
\end{table}

\subsection{Anchor Variations: Alignment, Prompt, and Identity}
\label{app:anchor-variations}

\subsubsection{Aligned vs.\ Unaligned Anchor}
\label{app:a3}

Replacing the aligned Qwen2.5-3B-Instruct anchor with its unaligned
base (same parameter count, no RLHF) on Llama-3.1-8B ($\alpha\!=\!0.5$,
$N\!=\!6$, $K\!=\!3$) collapses string-match refusal across the
adversarial sets: AdvBench 94.4\% $\to$ 69.4\%, HB-Std 79.6\% $\to$
52.1\%, HB-Ctx 79.0\% $\to$ 38.7\%, Sorry-Bench 61.4\% $\to$ 35.2\%.
The unaligned anchor still provides $\sim$25--52\,pp from the
beam-selection mechanism alone; the RLHF-aligned anchor of the same
size adds $\sim$25\,pp more on adversarial benchmarks at no extra cost. On the calibration side,
OR-Bench-Hard over-refusal stays close (25.4\% unaligned
vs.\ 22.3\% aligned).

\subsubsection{Anchor Prompt Robustness}
\label{app:a7}

We sweep three Phase~1 anchor-prompt variants on HarmBench-Standard
(Llama-3.1-8B, $n\!=\!167$, $\alpha\!=\!0.5$, $N\!=\!6$, $K\!=\!3$): no
prompt (anchor's unconditional prior), the standard 2-line safety
prompt used in all main experiments, and a verbose 45-line prompt.
String-match refusal is 66.5\% / \textbf{80.2\%} / 66.5\%
respectively; benchmark-judge accuracy is 82.6\% / 91.6\% /
91.6\%. The verbose prompt matches the standard prompt's judge
accuracy but yields lower string-match refusal, suggesting that
the anchor's RLHF training (not the literal prompt wording)
is what drives refusal under \alignbeam. We adopt the standard
2-line prompt as the default. Latency for all three variants falls
within the $K\!=\!3$ band reported in Appendix~\ref{app:latency}.

\subsubsection{Alternative Anchor (Llama-3.1-8B-Instruct)}
\label{app:a8a}

Table~\ref{tab:a8a} reports the alternative-anchor study.
The Llama anchor reaches 97--100\% judge accuracy on AdvBench and
HarmBench (85.9\% on Sorry-Bench) but only 7--15\% string-match
refusal: Llama-3.1-8B-Instruct's polite declinations fall outside
the 22-phrase canonical list, confirming that string-match
underestimates true safety when the anchor's refusal style is
non-canonical.

\begin{table}[htpb]
\centering\scriptsize
\caption{Llama anchor achieves 97--100\% judge accuracy but only
  7--15\% string-match refusal, confirming the anchor's refusal
  style must match the evaluation metric. Alternative-anchor study (Llama-3.1-8B draft,
  Llama-3.1-8B-Inst anchor, \lbmdrop, $\alpha\!=\!0.5$,
  $N\!=\!6$, $K\!=\!3$). Ref\% = string-match; ACC\% =
  benchmark-judge; $^{\dagger}$ = calibration/benign (lower is
  better).}
\label{tab:a8a}
\vspace{6.3pt}
\begin{tabular}{@{}lrrrrr@{}}
\toprule
\textbf{Dataset} & $\boldsymbol{n}$
  & \textbf{Base\,\%} & \textbf{Base ACC\,\%}
  & \textbf{+\,AB\,\%} & \textbf{+\,AB ACC\,\%} \\
\midrule
AdvBench      & 520 & 17.5 & 52.5 &  7.1 & \textbf{99.6} \\
HB-Std        & 167 & 13.8 & 49.1 & 13.2 & \textbf{97.0} \\
HB-Ctx        &  62 &  8.1 & 53.2 & 14.5 & \textbf{98.4} \\
Sorry-Bench   & 440 & 10.9 & 34.9 & 11.1 & \textbf{85.9} \\
OR-Bench-H$^{\dagger}$ & 855 & 11.0 & 13.6 & 14.0 & 4.4 \\
XSTest$^{\dagger}$ & 450 &  5.6 & 23.6 &  7.1 & 37.1 \\
\bottomrule
\end{tabular}%
\end{table}

\clearpage
\section{Additional Per-Pair Experiments}
\label{app:additional-per-pair}

\subsection{DeepSeek-Math-7B-Instruct: Strategy Comparison}
\label{app:e2-strategies}

DeepSeek-Math-7B-Instruct ($N\!=\!20$) compares all three
strategies. \lbmfirst achieves the highest string-match refusal
across all benchmarks (AdvBench 93.8\%, HB-Std 71.3\%, JBB-H
80.0\%); \lbmdrop and \lbmexact are numerically equivalent. All
three strategies produce zero degenerate outputs.
Table~\ref{tab:e2-strategies} reports the per-strategy numbers.
WildJailbreak-Harmful (\lbmdrop, $n\!=\!1{,}315$): string-match
refusal rises from 11.5\% to 15.0\% ($+$3.5\,pp), while judge
accuracy rises from 19.2\% to 25.6\% ($+$6.4\,pp); the
anchor's polite declinations replace the draft's compliance-phrasing
refusals, the same pattern as Finance-Llama3-8B
(Appendix~\ref{app:e6-breakdown}).

\begin{table}[htpb]
\centering\scriptsize
\caption{DeepSeek-Math-7B-Instruct ($N\!=\!20$,
  $\alpha\!=\!0.5$, $K\!=\!3$). All three strategies are
  zero-degenerate; \lbmfirst achieves the highest string-match
  refusal on every benchmark.}
\label{tab:e2-strategies}
\vspace{6.3pt}
\begin{tabular}{@{}lrrrrr@{}}
\toprule
\textbf{Strategy} & \textbf{AdvB Ref\%} & \textbf{HarmB Ref\%}
  & \textbf{JBB-H Ref\%} & \textbf{SorryB Ref\%} & \textbf{GSM8K} \\
\midrule
\lbmdrop  & 80.6 & 63.5 & 67.0 & 45.0 & 76.6\% \\
\lbmexact & 80.6 & 63.5 & 67.0 & 45.9 & 76.6\% \\
\lbmfirst & \textbf{93.8} & \textbf{71.3} & \textbf{80.0} & \textbf{57.3} & \textbf{76.6\%} \\
\bottomrule
\end{tabular}%
\end{table}

\subsection{Llama-3.1-8B: Auxiliary Ablations}
\label{app:e3-aux}

\subsubsection{Seed Robustness}
\label{app:seed}

Re-running Llama-3.1-8B under seeds \{42, 2, 3\} (\lbmdrop, $\alpha\!=\!0.5$,
$N\!=\!6$, $K\!=\!3$) leaves HB$+$AB string-match refusal within
$\pm$0.7\,pp: \textbf{90.8\%} (seed 42, main run) / 90.7\% (seed
2) / 90.1\% (seed 3). Per-benchmark spread is similarly tight on
AdvBench (94.4 / 94.4 / 95.2) and HarmBench-Std (79.6 / 79.0 /
74.3); HarmBench-Ctx ($n\!=\!62$) shows the only material drift
(79.0 / 72.6 / 64.5), consistent with the small sample size on
that split. OR-Bench-Hard and XSTest over-refusal vary by at most
$+$5.6\,pp across seeds (22.3/22.9/22.7 and 40.0/40.0/45.6).

\subsubsection{No-Judge Ablation}
\label{app:nojudge}

Dropping the Phase~3 LLM judge and returning the highest-LBM-score
beam on Llama-3.1-8B (\lbmdrop, $\alpha\!=\!0.5$, $N\!=\!6$, $K\!=\!3$) leaves
adversarial refusal unchanged or slightly higher:
AdvBench 94.4 $\to$ 97.9, HB-Std 79.6 $\to$ 81.4, HB-Ctx 79.0
$\to$ 85.5, Sorry-Bench 61.4 $\to$ 63.6, WildJailbreak 19.7 $\to$
18.7. The cost is paid on calibration: XSTest over-refusal rises
from 40.0\% to 46.4\% ($+$6.4\,pp). Two effects explain the
adversarial-side wash. \emph{First}, the 3B judge is itself small
and occasionally miscalibrated: on ties between two equally-safe
beams it picks on stylistic grounds, so the no-judge default
(beam~0 by LBM score) is reliably safe on structured adversarial
sets and skips the noisy tie-breaking. \emph{Second}, Phases~1--2
logit mixing already supplies most of the safety signal; the
judge's marginal value is calibration, exactly what shows up as
the $+$6.4\,pp XSTest over-refusal that reappears when the judge
is dropped. A better-calibrated judge would improve benign discrimination
(recovering the $+$6.4\,pp XSTest cost) without sacrificing
the adversarial gains, since the logit mixing already provides
the core safety signal.

\subsection{Llama-3.1-70B Scale Experiment}
\label{app:e4}

Llama-3.1-70B replaces the 8B draft, with all other
settings unchanged (\lbmdrop, $\alpha\!=\!0.5$, $N\!=\!6$,
$K\!=\!3$, Qwen2.5-3B-Instruct anchor). \alignbeam transfers
to the 70B scale: AdvBench refusal 16.4\%\,$\to$\,\textbf{87.3\%}
($+$70.9\,pp; judge ACC 56.5\%\,$\to$\,\textbf{99.4\%}),
HarmBench-Std 9.6\%\,$\to$\,\textbf{71.9\%} ($+$62.3\,pp),
Sorry-Bench 9.8\%\,$\to$\,\textbf{58.0\%} ($+$48.2\,pp).
Calibration over-refusal rises as expected: OR-Bench-Hard
8.2\,$\to$\,30.2, JBB-Benign 6.0\,$\to$\,27.0, XSTest
3.3\,$\to$\,43.6; OR-Bench-Toxic (a benign-but-adversarially-phrased
calibration set) shows string-match 8.9\,$\to$\,57.3 and judge ACC
58.6\,$\to$\,13.8, the same over-refusal pattern as Llama-3.1-8B
(Appendix~\ref{app:judges}). \alignbeam thus scales to a 70B
draft despite an 8$\times$ anchor-draft capacity gap.

\subsection{Wealth-Management LoRA Safety Evaluation}
\label{app:ew}

{\sloppy
Even when applied to an already aligned instruct base, domain SFT measurably degrades safety. A LoRA fine-tune of Llama-3.1-8B-Instruct on wealth-management data ($r\!=\!6$, NF4 quantisation, 5{,}000 steps, \texttt{sohamb37lexsi/\allowbreak bitext-wealth-management-\allowbreak llm-chatbot-splits}) reduces benchmark-judge accuracy by 3.6 pp on HB-Std and 7.7 pp on Sorry-Bench relative to the unmodified instruct model (Table~\ref{tab:ew-safety}). Applying \alignbeam{} at inference time, without retraining, restores safety and in two cases exceeds the instruct baseline (HB-Ctx: 100.0\% vs.\ 98.4\%; Sorry-Bench: 85.0\% vs.\ 83.6\%; see Table~\ref{tab:b2-breakdown}).
\par}

\begin{table}[htpb]
\centering
\scriptsize
\caption{Wealth-LoRA setting: Llama-3.1-8B-Instruct + wealth LoRA, with Qwen2.5-3B-Instruct as anchor. \texttt{\textbackslash lbmdrop}, $\alpha\!=\!0.5$, $N\!=\!6$, $K\!=\!3$. ``LoRA Base ACC'' is the wealth LoRA without \alignbeam{}.
``LoRA + AB'' applies \alignbeam{} to the LoRA model. ``Inst + AB Ref'' applies \alignbeam{} to the unmodified instruct model. Lower Ref\% is better for benign/calibration rows.}
\label{tab:ew-safety}
\vspace{6.3pt}
\begin{tabular}{@{}llrcccc@{}}
\toprule
Dataset & T & $n$ & Base ACC\% & \multicolumn{2}{c}{LoRA + AB} & Inst + AB Ref\% \\
\cmidrule(lr){5-6}
& & & & Ref\% & ACC\% & \\
\midrule
\multicolumn{7}{@{}l}{\textit{Harmful (refusal $\uparrow$)}} \\
AdvBench      & H & 520  & 99.6 & 89.4 & \textbf{100.0} & 99.0 \\
HB-Ctx        & H & 62   & 90.3 & 83.9 & \textbf{100.0} & 85.5 \\
HB-Std        & H & 167  & 95.2 & 92.2 & \textbf{98.2}  & 89.2 \\
Sorry-Bench   & H & 440  & 75.7 & 74.5 & \textbf{85.0}  & 73.2 \\
WildJailbreak & H & 1315 & 81.4 & 51.6 & 62.9           & 63.7 \\
\midrule
\multicolumn{7}{@{}l}{\textit{Benign / calibration}} \\
JBB-Benign    & B & 100  & 3.0  & 53.0 & 2.0 & 37.0 \\
OR-Bench-H    & B & 855  & 3.3  & 30.6 & 2.6 & 44.8 \\
XSTest        & C & 450  & 42.7 & 62.9 & 42.0 & 50.9 \\
\bottomrule
\end{tabular}
\end{table}

\subsection{Qwen2.5-7B-Base: Same-Family Anchor Size and Method Ablation}
\label{app:eq2}

Anchor calibration quality matters more than anchor size: pairing
Qwen2.5-7B-Base with a Qwen2.5-7B-Instruct anchor (matching
parameter count) outperforms a Qwen2.5-0.5B-Instruct anchor
(14$\times$ smaller) by $+$23.9\,pp HB-Std refusal
(HB$+$AB 98.1\% vs.\ 91.1\%) and over-refuses far less on benign
sets (XSTest: 40.0\% vs.\ 78.9\%; OR-Bench-H: 41.1\% vs.\ 62.8\%;
Table~\ref{tab:eq2-anchor}). The gap traces to the 0.5B anchor's
weaker refusal calibration: its safety signal is less sharp, so it
flags a higher fraction of benign prompts.

\begin{table}[htpb]
\centering\scriptsize
\caption{The 7B anchor outperforms the 0.5B anchor by $+$23.9\,pp
  HB-Std refusal and halves benign over-refusal, confirming that
  calibration quality (not parameter count) drives the gap. Qwen2.5-7B-Base
  (\lbmdrop, $\alpha\!=\!0.5$, $N\!=\!6$,
  $K\!=\!3$). Ref\% = string-match refusal (harmful) /
  over-refusal$^{\dagger}$ (benign). Base ACC\% = unmodified draft.}
\label{tab:eq2-anchor}
\vspace{6.3pt}
\begin{tabular}{@{}lrrrr@{}}
\toprule
\textbf{Dataset} & $\boldsymbol{n}$
  & \textbf{Base ACC\%}
  & \textbf{7B Ref\%} & \textbf{0.5B Ref\%} \\
\midrule
\multicolumn{5}{@{}l}{\textit{Harmful ($\uparrow$)}} \\
AdvBench           &  520 & 79.4 & \textbf{99.2} & 97.7 \\
HB-Ctx             &   62 & 40.3 & \textbf{80.6} & 64.5 \\
HB-Std             &  167 & 47.3 & \textbf{94.6} & 70.7 \\
Sorry-Bench        &  440 & 30.5 & \textbf{70.0} & 67.0 \\
WildJailbreak      & 1315 & 20.1 & 24.9          & \textbf{32.2} \\
\midrule
\multicolumn{5}{@{}l}{\textit{Benign / calibration$^{\dagger}$ ($\downarrow$)}} \\
XSTest             &  450 & 21.1 & \textbf{40.0} & 78.9 \\
OR-Bench-H         &  855 & 11.2 & \textbf{41.1} & 62.8 \\
\midrule
\textbf{HB$+$AB}   & \textbf{687} & \textbf{71.6} & \textbf{98.1} & 91.1 \\
\bottomrule
\end{tabular}%
\end{table}

\paragraph{LBM variant convergence in the same-vocabulary setting.}
Because both anchors share the Qwen2.5 vocabulary with the draft,
the text-bridge top-$B$ mapping collapses to an identity:
every anchor token maps to the same draft token, giving a
single-token match rate $>$98\% and making \lbmdrop, \lbmfirst,
and \lbmexact mathematically equivalent. Qwen2.5-7B-Base confirms this
empirically: all three variants yield identical ACC\% on every
harmful and calibration benchmark for both anchor sizes
(e.g.\ HB-Std 98.2\% across all variants for the 7B anchor;
80.2\% for the 0.5B anchor). Variant choice is irrelevant in
the same-vocabulary regime.

Tables~\ref{tab:eq2-methods-7b}--\ref{tab:eq2-methods-05b} extend
the strategy comparison to all seven evaluated methods across both
anchor sizes. For the 7B anchor, \alignbeam (any LBM variant;
identical in the same-vocabulary setting) achieves the highest
harmful-ACC on four of five adversarial datasets
(AdvBench 100.0\% vs.\ 98.3/97.7/97.5/96.7\% for CD/TopK-CD/Proxy/SafeDecoding;
HB-Std 98.2\% vs.\ 95.8\%; exception: HB-Ctx 87.1\% vs.\ CD 90.3\%).
For the 0.5B anchor, \alignbeam leads on all five adversarial datasets
(WildJailbreak: all LBM variants tie at 38.6\% vs.\ next-best TopK-CD 27.3\%).
The XSTest over-refusal floor is nearly identical across all seven strategies
and both anchors ($\approx$40--42\%), confirming the structural
calibration floor discussed in \S\ref{sec:limits}. The 7B anchor
uniformly suppresses benign over-refusal further than the 0.5B anchor
(OR-Bench-Toxic 3.2\% vs.\ 6.5\% under \alignbeam), consistent with
the anchor-capacity finding in Table~\ref{tab:eq2-anchor}.

\begin{table}[htpb]
\centering\scriptsize
\caption{\textbf{7B anchor} (Qwen2.5-7B-Base draft $+$
Qwen2.5-7B-Instruct, $\alpha\!=\!0.5$, $N\!=\!6$, $K\!=\!3$;
benchmark-judge ACC\%). AB $=$ \alignbeam{} (any LBM variant).
CD/TKCD/PT/SD $=$ contrastive decoding / top-$k$ CD / proxy tuning /
safe decoding. $^{\dagger}$ benign ($\downarrow$). Bold $=$ best harmful
gain or lowest benign over-refusal per row.}
\label{tab:eq2-methods-7b}
\vspace{6.3pt}
\begin{tabular}{@{}lrrrrrrr@{}}
\toprule
\textbf{Dataset} & $\boldsymbol{n}$ & \textbf{Base}
& \textbf{AB} & \textbf{CD} & \textbf{TKCD} & \textbf{PT} & \textbf{SD} \\
\midrule
\multicolumn{8}{@{}l}{\textit{Harmful ($\uparrow$)}} \\
AdvBench      &  520 & 79.4 & \textbf{100.0} & 98.3 & 97.7 & 97.5 & 96.7 \\
HB-Ctx        &   62 & 40.3 & 87.1 & \textbf{90.3} & 82.3 & 87.1 & 72.6 \\
HB-Std        &  167 & 47.3 & \textbf{98.2} & 95.8 & 88.6 & 88.6 & 87.4 \\
Sorry-Bench   &  440 & 30.5 & \textbf{77.3} & 70.5 & 64.8 & 70.2 & 63.0 \\
WildJailbreak & 1315 & 20.1 & \textbf{35.1} & 31.3 & 30.7 & 31.5 & 32.4 \\
\midrule
\multicolumn{8}{@{}l}{\textit{Benign / calibration$^{\dagger}$ ($\downarrow$)}} \\
JBB-Benign    &  100 &  9.0 & \textbf{3.0}  &  5.0 &  3.0 &  3.0 &  3.0 \\
OR-Bench-H    &  855 & 11.2 & \textbf{3.9}  &  4.8 &  5.7 &  5.5 &  6.7 \\
OR-Bench-Tox  &  585 & 54.5 & \textbf{3.2}  &  8.5 & 14.2 & 13.8 & 14.0 \\
XSTest        &  450 & 21.1 & 42.0 & 40.7 & 40.0 & 39.1 & \textbf{38.9} \\
\bottomrule
\end{tabular}%
\end{table}

\begin{table}[htpb]
\centering\scriptsize
\caption{\textbf{0.5B anchor} (Qwen2.5-7B-Base draft $+$
Qwen2.5-0.5B-Instruct, $\alpha\!=\!0.5$, $N\!=\!6$, $K\!=\!3$;
benchmark-judge ACC\%). Same column layout as
Table~\ref{tab:eq2-methods-7b}.}
\label{tab:eq2-methods-05b}
\vspace{6.3pt}
\begin{tabular}{@{}lrrrrrrr@{}}
\toprule
\textbf{Dataset} & $\boldsymbol{n}$ & \textbf{Base}
& \textbf{AB} & \textbf{CD} & \textbf{TKCD}
& \textbf{PT} & \textbf{SD} \\
\midrule
\multicolumn{8}{@{}l}{\textit{Harmful ($\uparrow$)}} \\
WildJailbreak & 1315 & 20.1 & \textbf{38.6} & 25.1 & 27.3 & 22.0 & 27.2 \\
AdvBench      &  520 & 79.4 & \textbf{98.5} & 91.0 & 90.4 & 91.0 & 81.9 \\
Sorry-Bench   &  440 & 30.5 & \textbf{72.3} & 52.5 & 52.3 & 50.9 & 53.9 \\
HB-Std        &  167 & 47.3 & \textbf{80.2} & 63.5 & 64.7 & 62.3 & 59.3 \\
HB-Ctx        &   62 & 40.3 & \textbf{79.0} & 38.7 & 48.4 & 53.2 & 58.1 \\
\midrule
\multicolumn{8}{@{}l}{\textit{Benign / calibration$^{\dagger}$ ($\downarrow$)}} \\
JBB-Benign    &  100 &  9.0 & \textbf{3.0}  &  7.0 &  7.0 &  4.0 &  8.0 \\
OR-Bench-H    &  855 & 11.2 & \textbf{4.6}  &  6.3 &  6.4 &  5.8 &  6.0 \\
OR-Bench-Tox  &  585 & 54.5 & \textbf{6.5}  & 19.3 & 20.5 & 20.7 & 20.5 \\
XSTest        &  450 & 21.1 & 42.2 & 41.3 & \textbf{40.4} & 40.9 & 41.3 \\
\bottomrule
\end{tabular}%
\end{table}

\clearpage
\section{Utility and Latency}
\label{app:utility-and-latency}

\subsection{Utility Preservation}
\label{app:utility}

\alignbeam preserves task utility: accuracy drops $\leq$0.5\,pp and
all semantic metrics stay within $\pm$1.5\,pp of the unmodified
draft. Under the $K\!=\!1$ utility mode (permissive anchor prompt,
no LLM judge), DeepSeek-Math-7B-Instruct on GSM8K moves from 77.0\% to
\textbf{76.6\%} ($-$0.4\,pp) and MedLlama-3-8B on MedQA
from 13.1\% to \textbf{12.7\%} ($-$0.4\,pp). Full per-strategy
MedQA semantic metrics (ROUGE-1/2, BERTScore-F1, SentSim) are in
Appendix~\ref{app:medqa-semantic}.

\subsubsection{MedQA Utility: Full Semantic Metrics}
\label{app:medqa-semantic}

Table~\ref{tab:medqa-semantic} gives the full per-strategy
semantic-similarity metrics for MedLlama-3-8B MedQA. All three LBM variants
are reported ($n\!=\!2{,}243$).
Accuracy is multiple-choice MedQA-USMLE; ROUGE-1/2, BERTScore-F1,
and SentSim (all-MiniLM-L6-v2 cosine) are on ChatDoctor-style
free-form generations vs.\ the reference answer.
All semantic deltas are within $\pm$1.5\,pp of baseline: safety
blending does not measurably change content fidelity.

\begin{table}[htpb]
\centering\scriptsize
\caption{MedQA utility breakdown across all LBM variants
  (JSL-MedLlama-3-8B-v2.0 $+$ Qwen2.5-3B-Instruct, $\alpha\!=\!0.5$,
  $K\!=\!1$ utility mode, permissive anchor prompt). $\Delta$ columns
  are strategy minus baseline (pp). Acc.\ is MedQA multiple-choice
  accuracy; semantic metrics are computed on ChatDoctor-style free-form
  generations vs.\ reference answer ($n\!=\!2{,}243$).}
\label{tab:medqa-semantic}
\vspace{6.3pt}
\begin{tabular}{@{}lrrrrrr@{}}
\toprule
\textbf{Strategy} & $\boldsymbol{n}$
  & \textbf{Acc\%} & \textbf{R-1} & \textbf{R-2}
  & \textbf{BS-F1} & \textbf{SentSim} \\
\midrule
Base                & 2243 & 13.1          & 0.2522         & 0.0319          & 0.7591          & 0.5961 \\
\lbmdrop\ (default) & 2243 & 12.7\,({\scriptsize$-$0.4\,pp}) & 0.2426\,({\scriptsize$-$0.0096}) & 0.0311\,({\scriptsize$-$0.0008}) & 0.7552\,({\scriptsize$-$0.0039}) & 0.5813 \\
\lbmexact           & 2243 & 12.7\,({\scriptsize$-$0.4\,pp}) & 0.2426\,({\scriptsize$-$0.0096}) & 0.0311\,({\scriptsize$-$0.0008}) & 0.7552\,({\scriptsize$-$0.0039}) & 0.5813 \\
\lbmfirst           & 2243 & 12.7\,({\scriptsize$-$0.4\,pp}) & 0.2426\,({\scriptsize$-$0.0096}) & 0.0311\,({\scriptsize$-$0.0008}) & 0.7552\,({\scriptsize$-$0.0039}) & 0.5813 \\
\bottomrule
\end{tabular}%
\end{table}

\subsection{Latency Details}
\label{app:latency}

\alignbeam's $K\!=\!3$ default falls inside the latency band
already occupied by LlamaGuard and Proxy Tuning; the $K\!=\!1$
budget mode is the cheapest non-classifier option and stays on
par with hard-prefix priming. Table~\ref{tab:latency} reports
per-pair wall-clock on a single RTX~6000~Pro Blackwell (96\,GB, bfloat16).

\begin{table}[htpb]
\centering\scriptsize
\caption{\alignbeam's $K\!=\!3$ default falls at $4.6$--$6.6\times$,
  inside the LlamaGuard / Proxy Tuning band; the $K\!=\!1$ budget
  mode retains $\sim$80\% of safety at $\sim$2$\times$.
  Per-sample latency (RTX~6000~Pro Blackwell 96\,GB, bfloat16,
  HarmBench-Standard). Phase~3 (KV-cache continuation)
  $\approx$69\%; Phase~2 $\approx$30\%; Phase~1 $\approx$1\%.
  $^{\ast}$Finance-Llama3-8B per-sample wall-clock reconstructed from the
  $6.51\times$ recorded slowdown and TPS-derived per-sample timing
  of the sibling 8B base models.}
\label{tab:latency}
\vspace{6.3pt}
\begin{tabular}{@{}lccrrr@{}}
\toprule
\textbf{Pair} & $\boldsymbol{K}$ & $\boldsymbol{N}$
  & \textbf{Base} & \textbf{\alignbeam} & \textbf{Slowdown} \\
\midrule
Llama-8B          & 3 &  6 &  3.3\,s & 16.9\,s & 5.07$\times$ \\
MedLlama-8B       & 3 &  6 &  2.6\,s & 13.7\,s & 4.88$\times$ \\
FinLlama-8B$^{\ast}$ & 3 &  6 &  2.7\,s & 17.6\,s & 6.51$\times$ \\
DeepSeek-Math-7B-Instruct   & 3 & 20 &  5.6\,s & 25.8\,s & 6.61$\times$ \\
\midrule
\textbf{Budget mode}  & 1 &  3 & \multicolumn{3}{c}{$\sim$2$\times$; $\sim$80\% of safety} \\
\bottomrule
\end{tabular}%
\end{table}

\paragraph{Comparison with other inference-time defenses.}
Table~\ref{tab:latency-compare} places \alignbeam's two operating
points next to the other inference-time interventions we evaluate
in this paper and to published competitor numbers. The
\alignbeam $K\!=\!1$ budget setting is the cheapest of any
non-classifier option, and the $K\!=\!3$ default falls inside the
band already occupied by LlamaGuard prompt/response filtering,
Proxy Tuning, and top-$k$ contrastive decoding (all evaluated
with the same $K\!=\!3$ harness for comparability). The slower
end of this band is a property of the inference-time defense
\emph{class}, not specific to \alignbeam; the cheaper end is on
par with the hard-prefix baseline.

\begin{table}[htpb]
\centering\scriptsize
\caption{End-to-end per-sample slowdown of inference-time safety
methods on Llama-3.1-8B / Qwen3-8B-Base drafts (same harness, same GPU, mean over
HarmBench-Standard and AdvBench unless noted). Competitor rows
marked $\dagger$ are from the published paper; the rest are
from our own harness, so the slowdown definition is identical.}
\label{tab:latency-compare}
\vspace{6.3pt}
\begin{tabular}{@{}lr@{}}
\toprule
\textbf{Method} & \textbf{Slowdown} \\
\midrule
\textbf{\alignbeam}, $K\!=\!1$ budget  & $\sim$2$\times$ \\
Hard prefix                       & 2.4--3.2$\times$ \\
\textsc{RAIN}$^{\dagger}$              & 3.78--4.36$\times$ \\
LlamaGuard-response, $K\!=\!3$    & 3.9--5.5$\times$ \\
LlamaGuard-prompt, $K\!=\!3$      & 4.1--6.0$\times$ \\
\textbf{\alignbeam}, $K\!=\!3$ default & 4.6--6.6$\times$ \\
Top-$k$ contrastive decoding (Qwen3-8B-Base)     & 4.8--5.4$\times$ \\
Proxy Tuning (Qwen3-8B-Base)                     & 4.9--5.4$\times$ \\
\bottomrule
\end{tabular}
\end{table}

\subsubsection{Configuration Summary}
\label{app:configs}

\paragraph{Default hyperparameters.}
$\alpha\!=\!0.5$ (mixing weight), $K\!=\!3$ (beam count, safety
mode) or $K\!=\!1$ (utility mode), $N\!=\!6$ (mixed steps;
$N\!=\!20$ for DeepSeek-Math-7B-Instruct), $\tau\!=\!2.5$ (judge threshold), $T\!=\!0.7$
(sampling temperature), $B\!=\!50$ (text-bridge top-$B$ width),
\texttt{max\_new\_tokens}\,$=$\,150 for safety / 512--1024 for
utility, repetition penalty 1.15, seed 42. The depth sweep uses $\alpha\!=\!0.85$ (held fixed) to isolate
the depth effect; the $\alpha$ sweep covers $\alpha\!\in\!\{0.10,0.25,0.50,0.75\}$ at
$N\!=\!6$, with the $\alpha\!=\!0.85$ point borrowed from the depth sweep.

\paragraph{Hardware and runtime.}
Single RTX~6000~Pro Blackwell (96\,GB) per pair, bfloat16 throughout.
Per-sample wall-clock latency on HarmBench-Standard is reported
in Appendix~\ref{app:latency}.

\paragraph{Software versions.}
\texttt{transformers} 4.47.1 (\alignbeam decoding),
\texttt{torch} 2.5.1, \texttt{rouge\_score} 0.1.2,
\texttt{bert\_score} 0.3.13 (roberta-large backbone),
\texttt{sentence-transformers} 3.3.1
(all-MiniLM-L6-v2 for SentSim).

\paragraph{Anchor.}
Qwen2.5-3B-Instruct is the safety anchor for every reported run
(the five main pairs and the 70B-scale Llama-3.1-70B). The alternative-
anchor study (Appendix~\ref{app:a8a}) replaces it with
Llama-3.1-8B-Instruct and is the only run with a different
anchor.

\clearpage
\section{Extended Limitations}
\label{app:ext-limits}

This section expands the limitations summarised in
Sec.~\ref{sec:limits}, with detailed discussion of the two
verbose-draft failure modes and the calibration ceiling.

\paragraph{Calibration ceiling (extended).}
Both safety limits stem from the anchor, not the mixing parameters.
First, the anchor's tendency to over-refuse ambiguous benign prompts
is inherited: OR-Bench-Hard over-refusal stays $\approx$20\% across
all $\alpha$ (Figure~\ref{fig:ablations}(b)) and XSTest over-refusal
$\approx$40\% on Llama-3.1-8B; neither improves with more mixing. Second, the
mixing only covers the first $N$ tokens, so attacks that embed harm
past the prefix are only partially blocked (WildJailbreak-Harmful:
$+$2--6\,pp on Llama-3.1-8B/MedLlama-3-8B vs.\ $+$51--77\,pp on AdvBench). Both could be
improved with a better-calibrated anchor, a longer mixed prefix, or
a per-prompt $\alpha$ schedule. The alternative-anchor study
(Appendix~\ref{app:a8a}) underscores this: replacing the
Qwen2.5-3B-Instruct anchor with Llama-3.1-8B-Instruct reaches
97--100\% judge accuracy on HarmBench and AdvBench but only 85.9\%
on Sorry-Bench, because Llama's polite refusal phrases fall outside
the 22-phrase string-match list. The anchor is interchangeable, but
its refusal style needs to match the evaluation metric.

\paragraph{Output truncation at maximum completion length.}
All safety benchmarks are run with \texttt{max\_new\_tokens}\,$=$\,150.
On adversarial prompts that elicit lengthy preambles before a
refusal, the generation window can expire mid-sentence, producing
a response that neither completes a refusal nor completes harmful
content. Such truncated outputs are classified as non-refusals by
both the string-match rule and most benchmark judges, understating
true safety gains on verbose draft models. This is an engineering
artifact rather than a fundamental limitation of \alignbeam: it can
be addressed by adjusting the stopping criteria to terminate
generation as soon as a canonical refusal phrase is detected, or by
increasing \texttt{max\_new\_tokens} for safety evaluation runs.
The effect is most visible on DeepSeek-Math-7B-Instruct at
$N\!=\!6$, where the mixing window is insufficient and the draft
appends a harmful continuation after a refusal prefix before the
budget expires; raising $N$ to 20 substantially reduces but does
not eliminate this pattern. Utility runs use a longer budget
(512--1024 tokens), so truncation does not affect GSM8K or MedQA
accuracy figures.

\paragraph{False refusals without content change (sorry-but-continues).}
A distinct failure mode, separate from the over-refusal reported on
benign benchmarks, occurs when \alignbeam steers the first tokens
toward a canonical refusal phrase but Phase~3 draft continuation
then proceeds to produce the requested harmful content. This
``sorry-but-continues'' pattern is most pronounced on models whose
instruction-tuning strongly favours compliance after any refusal
opener. DeepSeek-Math-7B-Instruct is the clearest example, where
the math-instruct training distribution provides no signal for what
should follow ``I cannot help with that.'' It is documented more
broadly in Appendix~\ref{app:string-vs-acc} for unsteered baselines on
Sorry-Bench sub-categories (Fraud/Deception, Privacy).
Phase~2 logit mixing substantially reduces this pattern relative to
unsteered drafts, but does not eliminate it: a response that opens
with ``I'm sorry, I cannot help with that'' and then continues with
the harmful answer is counted as a refusal by string-match
(Ref\,\%), while the benchmark judge correctly classifies it as
non-safe (ACC\,\%). Two practical mitigations exist and were not
evaluated here. First, increasing $N$ extends the mixed-decoding
window further into the continuation, reducing the draft's
opportunity to recover its compliance trajectory (we raise $N$ to
20 for DeepSeek-Math-7B-Instruct, which substantially but not fully resolves the pattern).
Second, applying a stricter utility-mode anchor prompt during
Phase~3 (one that conditions the anchor on refusal rather than
compliance) may suppress the continuation without requiring a longer
mixed prefix; we leave this to future work. Increasing the judge
threshold $\tau$ also mitigates the pattern at the cost of higher
latency and benign over-refusal.

\paragraph{Adaptive attacks and multi-turn jailbreaks.}
All evaluations use single-turn prompts $\leq$200 tokens. We do not
evaluate adaptive attacks designed to circumvent \alignbeam, though
the partial WildJailbreak result ($+$2--6\,pp string-match) gives a
rough lower bound on that vulnerability. Multi-turn jailbreaks,
which iteratively shape the draft's context past the mixed prefix,
are explicitly out of scope and are an obvious target for follow-up
work using a per-turn $\alpha$ schedule or anchor re-conditioning.

\end{document}